\documentclass{article}

\usepackage[preprint]{neurips_2026}

\usepackage[utf8]{inputenc} %
\usepackage[T1]{fontenc}    %
\usepackage{hyperref}       %
\usepackage{url}            %
\usepackage{booktabs}       %
\usepackage{amsfonts}       %
\usepackage{amsmath}        %
\usepackage{nicefrac}       %
\usepackage{microtype}      %
\usepackage{xcolor}         %
\usepackage{graphicx}       %
\makeatletter
\@ifundefined{KV@Gin@alt}{\define@key{Gin}{alt}{}}{}
\makeatother
\usepackage{multirow}       %
\usepackage{enumitem}       %
\usepackage{float}          %
\usepackage{rotating}       %
\usepackage[breakable]{tcolorbox} %

\newcommand{\shieldNotes}{1{,}381}

\newcommand{\shieldSpans}{10{,}229}
\newcommand{\shieldTokens}{402{,}647}
\newcommand{\shieldWords}{232{,}090}

\newcommand{\shieldNumCats}{9}
\newcommand{\shieldNumCatsWord}{nine}
\newcommand{\shieldCatList}{\texttt{AGE}, \texttt{DATE}, \texttt{DOCTOR}, \texttt{HOSPITAL}, \texttt{ID}, \texttt{LOCATION}, \texttt{PATIENT}, \texttt{PHONE}, and \texttt{WEB}}
\newcommand{\shieldSelectedNotes}{1{,}394}
\newcommand{\shieldDuplicates}{13}
\newcommand{\shieldNumLabelers}{12}
\newcommand{\shieldDateSpans}{3{,}547}
\newcommand{\shieldLocationSpans}{445}
\newcommand{\shieldWebSpans}{82}

\newcommand{\debDoctorR}{0.90}
\newcommand{\biomodernDoctorR}{0.87}
\newcommand{\debPatientR}{0.88}
\newcommand{\biomodernPatientR}{0.81}
\newcommand{\debHospR}{0.79}
\newcommand{\biomodernHospR}{0.72}
\newcommand{\debAgeR}{0.78}
\newcommand{\debLocR}{0.55}
\newcommand{\debWebR}{0.67}
\newcommand{\debDateR}{0.94}
\newcommand{\aimivOneHospRCI}{0.90 [0.88--0.93]}
\newcommand{\debHospRCI}{0.79 [0.75--0.82]}
\newcommand{\dateRecallLo}{0.94}
\newcommand{\dateRecallHi}{0.95}
\newcommand{\debHospItwobR}{0.39}
\newcommand{\debHospAimiR}{0.02}
\newcommand{\studentMicroP}{0.89}
\newcommand{\studentMicroR}{0.88}
\newcommand{\studentMacroP}{0.85}
\newcommand{\studentMacroR}{0.81}
\newcommand{\teacherMacroP}{0.91}
\newcommand{\teacherMacroR}{0.90}

\newcommand{\geminiFlashMacroR}{0.90}
\newcommand{\geminiProMacroR}{0.90}
\newcommand{\distAgeDelta}{-0.12}
\newcommand{\distAgeP}{<0.0005}
\newcommand{\distDateDelta}{-0.04}
\newcommand{\distDateP}{<0.0005}
\newcommand{\distDoctorDelta}{-0.06}
\newcommand{\distDoctorP}{<0.0005}
\newcommand{\distHospitalDelta}{-0.09}
\newcommand{\distHospitalP}{<0.0005}
\newcommand{\distIdDelta}{-0.05}
\newcommand{\distIdP}{=0.0040}
\newcommand{\distLocationDelta}{-0.11}
\newcommand{\distLocationP}{<0.0005}
\newcommand{\distPatientDelta}{-0.07}
\newcommand{\distPatientP}{<0.0005}
\newcommand{\distPhoneDelta}{-0.10}
\newcommand{\distPhoneP}{<0.0005}
\newcommand{\distWebDelta}{-0.15}
\newcommand{\distWebP}{=0.0215}
\newcommand{\aimiNotes}{46{,}313}
\newcommand{\aimiNumCats}{8}
\newcommand{\starrChars}{633 billion}

\newcommand{\starrTokens}{158 billion}
\newcommand{\starrTokensShort}{158B}
\newcommand{\teacherLabelNotes}{$\sim$13{,}000}
\newcommand{\teacherLabelNotesShort}{$\sim$13k}
\newcommand{\studentParams}{184M}
\newcommand{\gptossTotal}{117B}
\newcommand{\gptossActive}{5.1B}

\title{SHIELD: A Diverse Clinical Note Dataset and Distilled Small Language Models for Enterprise-Scale De-identification}

\author{%
  Jose D. Posada, Ph.D \thanks{Corresponding author: \texttt{jdposada@stanford.edu}} \\
  Technology \& Digital Solutions \\
  Stanford Medicine \\
  \And
  David Love M.Sc\\
  Technology \& Digital Solutions \\
  Stanford Medicine \\
  \And
  Somalee Datta, Ph.D \\
  Technology \& Digital Solutions \\
  Stanford Medicine \\
  \And
  Priya Desai \\
  Technology \& Digital Solutions \\
  Stanford Medicine \\
}

\begin{document}

\maketitle

\begin{abstract}
  De-identification of clinical text remains a prerequisite for the secondary use of electronic health records (EHRs). Existing public benchmarks, notably the i2b2 2006 and 2014 corpora, are over a decade old and lack the semantic and demographic diversity of modern clinical narratives. Large Language Models (LLMs) reach state-of-the-art zero-shot extraction, but their use at enterprise scale is limited by computational cost and by hospital data governance that restricts sending Protected Health Information (PHI) to cloud APIs. We introduce SHIELD (Synthetic Human-annotated Identifier-replaced Entries for Learning and De-identification), a diverse clinical note dataset containing \shieldNotes{} notes with \shieldSpans{} gold-standard PHI spans across \shieldNumCats{} categories, built using set-cover diversity sampling across demographic and document-type strata with human-in-the-loop adjudication. We evaluate four LLMs (two proprietary and two open-weight) to establish a performance ceiling on SHIELD, then show that a teacher-student distillation framework transfers these capabilities into locally deployable Small Language Models (SLMs). Our best distilled model reaches micro-averaged span-level precision of \studentMicroP{} and recall of \studentMicroR{} while running on standard workstation hardware. It trails its cloud teacher on per-category recall (a \teacherMacroR{} vs.\ \studentMacroR{} macro-averaged recall gap) but remains competitive given its substantially lower cost and on-premise deployability. Cross-dataset evaluation across corpora from different institutions shows that diversity-trained models generalize well on universal structured PHI categories, while institution-specific entities remain hard to transfer in both directions. This pattern suggests that optimal deployment may combine broad-coverage models with specialized models for high-volume, semi-structured note types. We publicly release the SHIELD dataset and the distilled DeBERTa v3 model to provide an accurate, cost-effective de-identification pipeline deployable entirely behind institutional firewalls.
\end{abstract}

\section{Introduction}

\paragraph{The data bottleneck in healthcare AI.}
The digitization of medical records has created large repositories of unstructured clinical text that are essential for data-driven medical research and for building clinical artificial intelligence (AI) \citep{kather2024llm}. These free-text narratives are laden with Protected Health Information (PHI). Privacy requirements under regulations such as the Health Insurance Portability and Accountability Act (HIPAA) and the General Data Protection Regulation (GDPR) require that PHI be de-identified before secondary use or multi-institutional sharing. A systematic review of 69 studies reported that machine-learning and hybrid approaches now dominate clinical de-identification research, yet progress remains tied to the availability of high-quality annotated corpora \citep{kovacevic_-identification_2024}. Scoping reviews of the broader literature reach the same conclusion, naming annotated data scarcity as the primary bottleneck limiting deployment at scale \citep{negash_-identification_2023}.

\paragraph{Limitations of existing approaches and benchmarks.}
Public benchmarks have driven methodological progress. The i2b2 2006 challenge provided the first shared evaluation for de-identification systems \citep{uzuner_evaluating_2007}, and the i2b2 2014 corpus established a widely adopted standard \citep{stubbs2015corpus}. Both datasets are now over a decade old and focus narrowly on particular phenotypes. On the methods side, rule-based systems suffer from low precision and over-redaction, and transformer-based models require costly local adaptation to each institution \citep{chambon_automated_2023}. LLM-based approaches such as DeID-GPT \citep{liu_deid-gpt_2023} offer zero-shot capabilities but introduce hallucination risks, high inference costs, and data-governance concerns when PHI must traverse cloud APIs. Efforts to scale de-identification to billions of clinical notes \citep{kocaman_beyond_2023} point to the need for systematic, multi-system evaluation frameworks \citep{heider_extensible_2024-2}.

\paragraph{Cross-institution generalization gap.}
An under-addressed challenge is the generalization gap: models trained on one institution's corpus often fail when applied elsewhere. \citet{yang_study_2019} report that a deep-learning de-identification model's F1 score dropped from 0.96 to 0.86 when evaluated across institutions. Studies at the US Department of Veterans Affairs \citep{eyre_evaluating_2025} and a multi-specialty UK hospital \citep{kuo_benchmarking_2025} report that transformer transferability remains fragile, and deployment experience with AnonCAT \citep{kraljevic_validating_2023} shows the operational difficulty of maintaining accuracy outside the training domain.

\paragraph{Contributions.}
To resolve these bottlenecks, we make four contributions:
\begin{itemize}
    \item We introduce \textbf{SHIELD} (Synthetic Human-annotated Identifier-replaced Entries for Learning and De-identification), a new diverse benchmark with human-in-the-loop annotation using LLM pre-annotation and adjudication across \shieldNumCats{} PHI categories.
    \item We present a knowledge-distillation framework that transfers the reasoning capabilities of a cloud LLM (Gemini 2.5 Flash) into Small Language Models (SLMs), specifically DeBERTa v3 \citep{he2021deberta} and BioClinical ModernBERT \citep{sounack_bioclinical_2025}, reaching micro-averaged span-level precision \studentMicroP{} and recall \studentMicroR{} at a fraction of the cost. This is consistent with recent evidence that SLMs are becoming the practical backbone of deployed AI systems \citep{belcak_small_2025, lee_targeted_2025}.
    \item We evaluate four LLMs and four transformer models, establishing a performance ceiling on SHIELD and measuring cross-dataset generalization across three datasets (SHIELD, i2b2 2014, AIMI).
    \item We publicly release the SHIELD dataset and the distilled DeBERTa v3 model to lower the barrier to entry for clinical institutions wishing to safely unlock their EHR data.\footnote{Access instructions available at \url{https://github.com/susom/shield_dataset}}
\end{itemize}

\section{Methods}

\subsection{Dataset construction and diversity sampling}

To construct a benchmark that reflects diverse patient populations and clinical scenarios, we bypassed random sampling. We developed a sampling pipeline applied to the Stanford Medicine Research Data Repository, STARR-OMOP \citep{datta2020newparadigmacceleratingclinical, callahan2023starr}, which contains over \starrTokens{} tokens of clinical text. We used a Set Cover Algorithm \citep{williamson2011design} for diversity sampling across six demographic and document axes: age, sex, race, ethnicity, note type, and note length. The algorithm iteratively selects notes that maximize coverage of under-represented strata, producing a compact corpus that spans the demographic range of the source warehouse.

\begin{figure}[t]
  \centering
  \includegraphics[width=0.6\textwidth,angle=0,alt={SHIELD annotation pipeline diagram: STARR-OMOP notes flow through diversity sampling, LLM pre-annotation, two-reviewer human labeling, and adjudication into the gold-standard corpus.}]{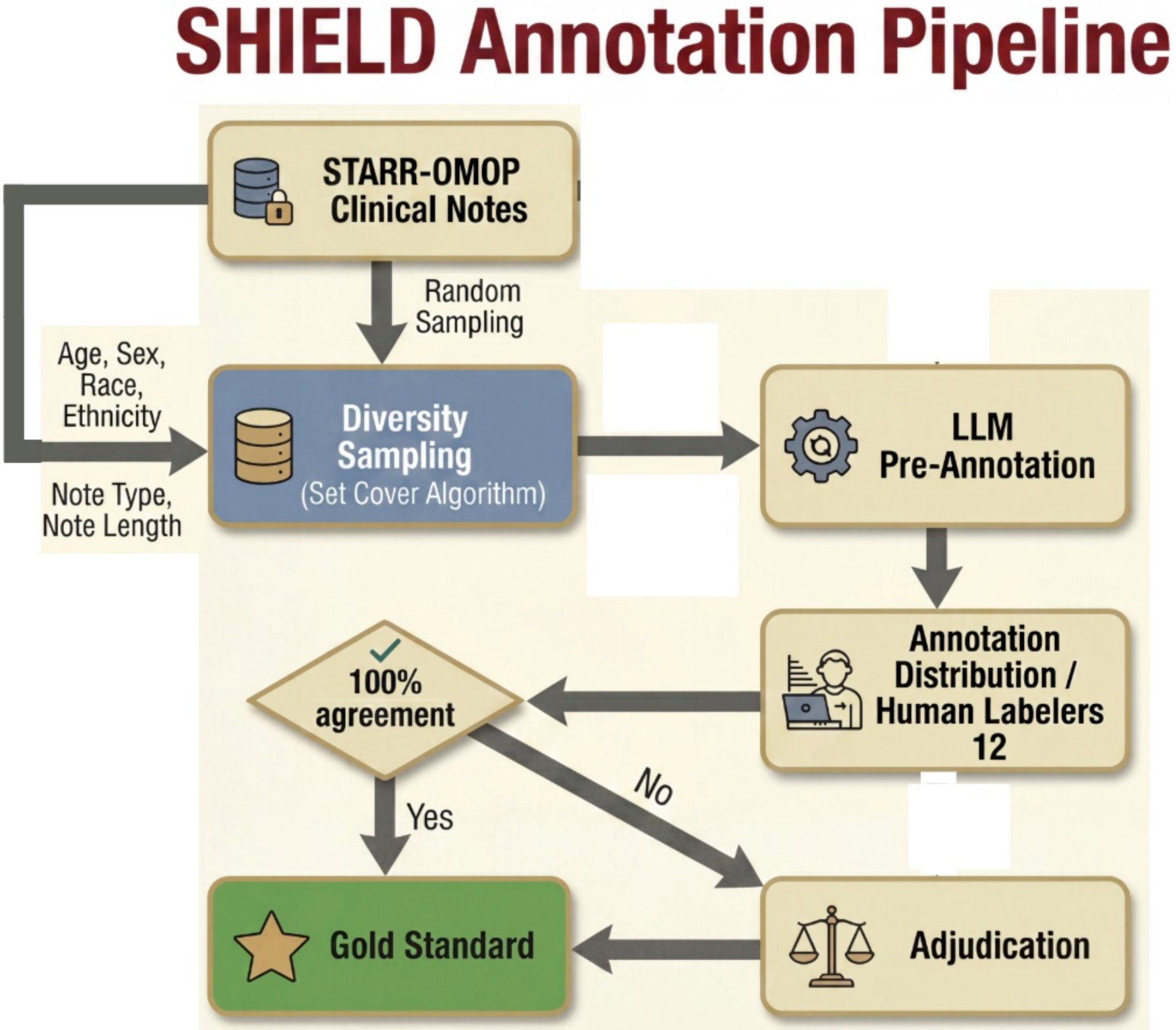}
  \caption{SHIELD annotation pipeline. Clinical notes are sampled from STARR-OMOP via diversity-optimized set cover, pre-annotated by an LLM, independently reviewed by two of 12 trained human labelers per note, and adjudicated to produce gold-standard PHI labels.}
  \label{fig:annotation-pipeline}
\end{figure}

The selected \shieldSelectedNotes{} notes underwent a human-in-the-loop annotation process (Figure~\ref{fig:annotation-pipeline}). Notes were first pre-annotated using an LLM to identify candidate PHI spans. These pre-annotations were distributed to a pool of \shieldNumLabelers{} trained human labelers, with each note independently reviewed by two annotators. Spans with 100\% inter-annotator agreement between the two reviewers were accepted directly into the gold standard; all disagreements were adjudicated by a senior annotator. After removing \shieldDuplicates{} exact-duplicate notes during the de-identification described in Section~\ref{sec:dataset-release}, the released corpus contains \shieldNotes{} notes with \shieldSpans{} gold-standard PHI spans across \shieldNumCats{} PHI categories: \shieldCatList{}. To enable cross-dataset benchmarking against i2b2 2014 \citep{stubbs2015corpus} and AIMI, a large-scale radiology report de-identification corpus developed by the AIMI Center at Stanford \citep{prakash_improving_2025} and made available for this study through a research collaboration, all annotations were mapped into a unified canonical label taxonomy (see Appendix~\ref{appendix:label-mappings} for the complete mapping definitions).

\subsection{Dataset release and surrogate replacement}
\label{sec:dataset-release}

The released SHIELD artifact is designed to protect patient privacy while preserving the linguistic structure required for meaningful evaluation of de-identification systems. We emphasize that the underlying narratives are \emph{real} STARR-OMOP clinical notes; the ``Synthetic'' in SHIELD refers only to the surrogate identifiers substituted for PHI, not to the clinical text itself, which retains its original distributional realism. Rather than redacting PHI spans, we replace each span with a type-appropriate cryptographic surrogate, yielding a corpus that reads as natural clinical text and supports end-to-end token- and span-level evaluation. Surrogates are generated by entity type. PATIENT and DOCTOR names are replaced with credible surrogate names of the same form. DATE values are shifted by a per-patient jitter of 3--90 days, derived deterministically from each patient's identifier via HMAC-SHA256 and applied uniformly to every date in that patient's notes, which preserves relative intervals between events while preventing calendar-based re-identification. ID and MRN values are replaced using format-preserving encryption, so each surrogate keeps the length and character structure of the original. PHONE and LOCATION values are replaced with credible random surrogates in a format-preserving form. AGE values are left unchanged, except that ages above 89 are capped at 89 where they appear in a note, following HIPAA's Safe Harbor treatment of ages over 89. Post-processing recomputes text hashes over the surrogate text, replaces patient identifiers with one-way hashes, and realigns each PHI span's character offsets to the surrogate-replaced output. This approach follows the precedent set by the i2b2 2014 release \citep{stubbs2015corpus} and recently applied to MC-MED \citep{kansal2025mcmed}. Per-category PHI distribution, note-length statistics, patient demographics, and the most common note types are reported in Appendix~\ref{appendix:release}.

\subsection{Model selection and distillation framework}
\label{sec:model-selection}

Our objective was to connect high-capability LLMs with firewall-friendly deployment. To establish a performance ceiling for current LLMs on de-identification, we first evaluated four models on the SHIELD gold standard in a zero-shot extraction setting. We tested two proprietary models, Gemini 2.5 Pro and Gemini 2.5 Flash (Google), and two open-weight models, OpenAI's GPT-OSS 120B (\gptossTotal{} total parameters, \gptossActive{} active; Mixture-of-Experts, Apache 2.0 license) \citep{openai_gptoss_2025} and Meta's Llama 4 Maverick (Mixture-of-Experts, Llama 4 Community License) \citep{llama4herd_2026}. All LLMs were prompted with the same extraction template targeting the \shieldNumCats{} canonical PHI categories (see Appendix~\ref{appendix:llm-prompt} for the full prompt). Gemini 2.5 Flash, which we adopt as the distillation teacher, is close to the top model (macro-averaged span-level recall \geminiFlashMacroR{}, vs.\ \geminiProMacroR{} for Gemini 2.5 Pro) while also offering the deployment economics (low cost, high throughput, strong instruction-following, and a large context window; Section~\ref{sec:distillation-results}) that make at-scale silver labeling practical.

\begin{figure}[t]
  \centering
  \includegraphics[width=0.6\textwidth,angle=-90,alt={Teacher-student distillation diagram: a small labeled sample creates prompts for Gemini 2.5 Flash, which labels approximately \teacherLabelNotes{} unlabeled notes used to train DeBERTa v3 and BioClinical ModernBERT students.}]{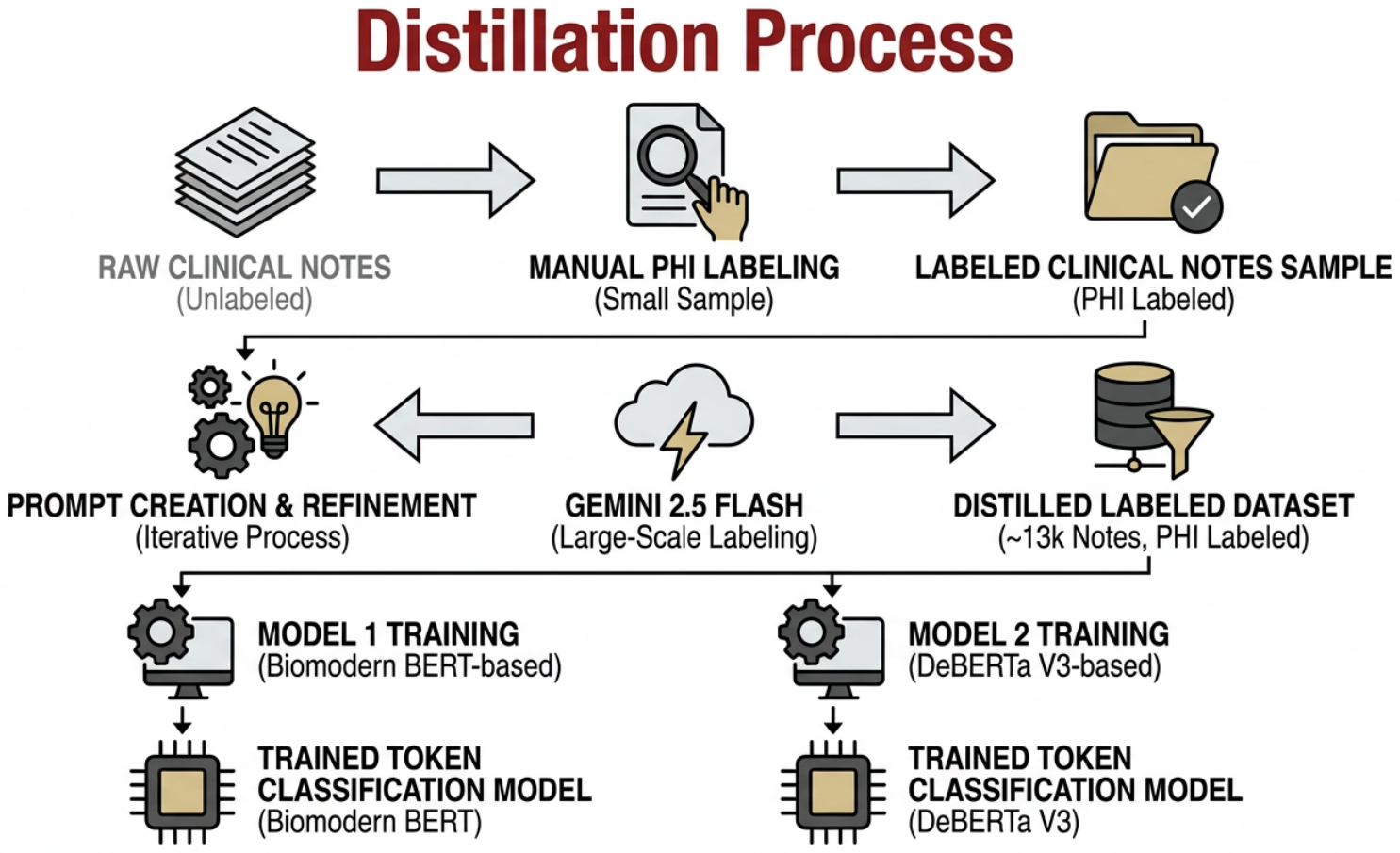}
  \caption{Teacher-student distillation process. A small labeled sample is used to create prompts for Gemini 2.5 Flash, which then labels \teacherLabelNotesShort{} notes. The distilled dataset trains DeBERTa v3 and BioClinical ModernBERT for local deployment.}
  \label{fig:distillation}
\end{figure}

We then constructed a three-stage teacher-student distillation pipeline (Figure~\ref{fig:distillation}):

\begin{enumerate}
    \item \textbf{Prompt creation:} A small subset of the SHIELD gold standard was used solely to iteratively craft and refine extraction prompts. These prompts were validated against the gold standard until satisfactory precision and recall were achieved on the calibration set. No SHIELD samples were used to train the student models.
    \item \textbf{Large-scale teacher labeling:} The refined prompts were applied with Gemini 2.5 Flash to approximately \teacherLabelNotes{} unlabeled clinical notes from the STARR-OMOP warehouse, generating silver-standard annotations at scale. Gemini 2.5 Flash was selected as the primary teacher due to its superior instruction-following, expansive context window, and state-of-the-art extraction accuracy.
    \item \textbf{Student model training:} The distilled silver-standard labels were used to train two SLM architectures via token-level cross-entropy over BIO tags: DeBERTa v3 \citep{he2021deberta} and BioClinical ModernBERT \citep{sounack_bioclinical_2025}. We selected these two architectures based on controlled comparisons showing that DeBERTa v3 offers better sample efficiency and benchmark performance, while ModernBERT offers advantages in long-context support and inference speed \citep{antoun_modernbert_2025}. No SHIELD gold-standard annotations were included in the training data, which preserves the entire SHIELD dataset as an independent evaluation set. Both architectures support efficient inference, running on CPUs or consumer-grade edge GPUs within a hospital's protected IT environment.
\end{enumerate}

We also benchmarked two prior-generation models, AIMI v1 and AIMI v2, developed by the AIMI Center at Stanford as part of their radiology report de-identification effort \citep{prakash_improving_2025}. Both models were trained on the AIMI dataset, a corpus of \aimiNotes{} radiology reports with gold-standard PHI annotations across \aimiNumCats{} categories. The AIMI dataset is not publicly available; access was provided for this study through a collaboration with the AIMI researchers. AIMI v1 and v2 are successive iterations of DeBERTa-based de-identification models trained on this domain-specific radiology data using the AIMI annotation pipeline. All four transformer models were evaluated on all three datasets (SHIELD, i2b2, AIMI) at both span-level and token-level granularity.

\subsection{Statistical analysis}
\label{sec:statistical-analysis}

Span-level evaluation uses an overlap-based matching criterion: a predicted span is counted as a true positive if at least 80\% of the gold span's character length is covered by the prediction; unmatched predictions are false positives and unmatched gold spans are false negatives.

To quantify uncertainty in performance estimates, especially for low-support PHI categories, we computed bootstrap 95\% confidence intervals (CIs) using document-level stratified resampling. For each evaluation, we resampled documents with replacement 2{,}000 times (seed\,=\,42), recomputed span-level precision and recall per category on each bootstrap sample, and derived CIs using the percentile method. To test whether differences between models are statistically significant, we performed paired bootstrap tests: on each resample, the difference in recall between two models was computed, and the two-sided $p$-value was estimated as the proportion of bootstrap samples where the observed difference changed sign. All $p$-values were Bonferroni-corrected for multiple comparisons across the 9 PHI categories ($\alpha = 0.05 / 9 \approx 0.0056$).

\section{Results}

\subsection{SHIELD dataset statistics}

\begin{figure}[t]
  \centering
  {\small\textbf{(a)}}\\[0.5em]
  {\small
  \begin{tabular}{lrrrrr}
    \toprule
    \textbf{Dataset} & \textbf{Notes} & \textbf{Tokens} & \textbf{Words} & \textbf{Vocab} & \textbf{PHI Spans} \\
    \midrule
    SHIELD & 1{,}381 & 402{,}647 & 232{,}090 & 39{,}762 & 10{,}229 \\
    I2B2 & 1{,}304 & 1{,}399{,}589 & 805{,}117 & 73{,}091 & 28{,}867 \\
    AIMI & 46{,}313 & 9{,}688{,}355 & 5{,}547{,}706 & 140{,}026 & 172{,}276 \\
    \bottomrule
  \end{tabular}}
  \par\vspace{1em}
  {\small\textbf{(b)}}\\[0.5em]
  \includegraphics[width=0.85\textwidth,alt={Bar chart of PHI category counts across SHIELD, i2b2, and AIMI on a log scale, showing per-category support for the unified label taxonomy.}]{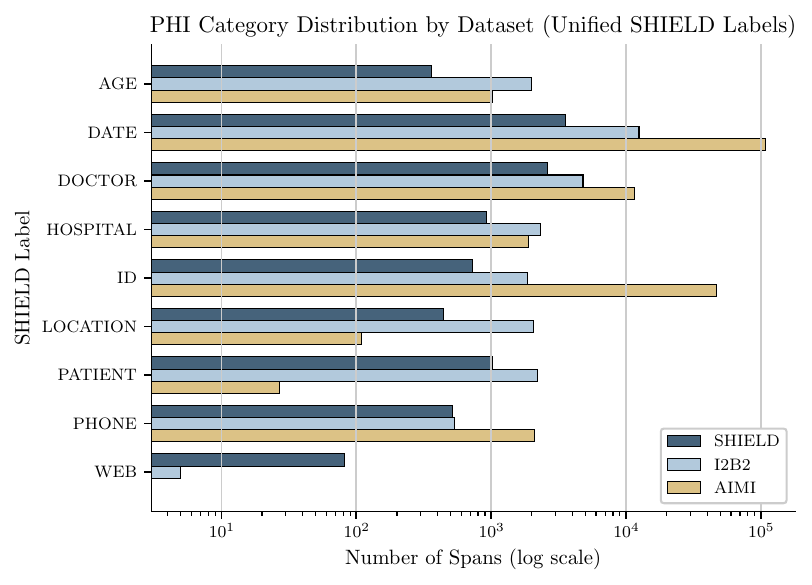}
  \caption{Dataset overview. (a)~Statistics for the three evaluation corpora. (b)~PHI category distribution across SHIELD, i2b2, and AIMI using the unified label taxonomy (log scale).}
  \label{fig:dataset-overview}
\end{figure}

Figure~\ref{fig:dataset-overview}(a) summarizes the three evaluation datasets. The final SHIELD dataset comprises \shieldNotes{} notes containing an estimated \shieldTokens{} tokens, \shieldWords{} words, and \shieldSpans{} gold-standard PHI spans across \shieldNumCats{} categories. While SHIELD is smaller in total note volume than AIMI (\aimiNotes{} notes) its diversity sampling strategy was designed to capture a broad range of clinical language across demographic and document-type strata. Figure~\ref{fig:dataset-overview}(b) shows the per-category PHI distribution across the three datasets.

\subsection{LLM benchmark on SHIELD}

\begin{figure}[t]
  \centering
  \includegraphics[width=\textwidth,alt={Radar plots of span-level precision and recall across \shieldNumCatsWord{} PHI categories for four large language models (Gemini 2.5 Pro, Gemini 2.5 Flash, Llama 4 Maverick, GPT-OSS 120B) evaluated zero-shot on SHIELD.}]{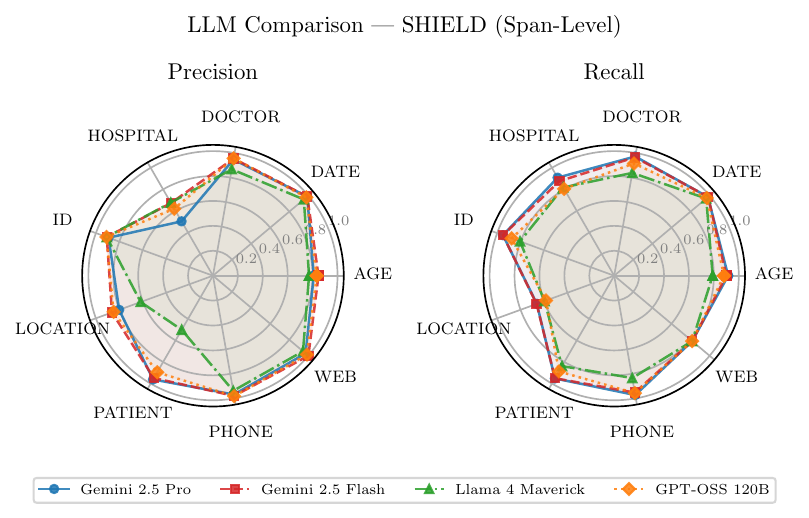}
  \caption{LLM benchmark on SHIELD. Span-level radar profiles showing precision (left) and recall (right) for four LLMs in zero-shot extraction. All models achieve strong coverage on structured categories (DATE, PHONE) but diverge on the recall of context-dependent entities (LOCATION, WEB), and on HOSPITAL precision (0.50--0.68) despite uniformly high HOSPITAL recall. Full numerical results are provided in Table~\ref{tab:llm-span} (Appendix~\ref{appendix:supplementary}).}
  \label{fig:llm-benchmark}
\end{figure}

Figure~\ref{fig:llm-benchmark} presents the span-level performance of four LLMs evaluated on SHIELD in a zero-shot extraction setting (full numerical results in Table~\ref{tab:llm-span}, Appendix~\ref{appendix:supplementary}). All four models perform well across most PHI categories, which supports LLMs as viable de-identification tools. Gemini 2.5 Pro and Gemini 2.5 Flash lead overall, with the highest macro-averaged recall (\geminiProMacroR{} and \geminiFlashMacroR{}, respectively); Gemini 2.5 Pro reaches the top AGE recall (0.91). The open-weight GPT-OSS 120B stays competitive on structured categories (DATE R\,=\,0.97, PHONE R\,=\,0.96), while Llama 4 Maverick trails on recall for DOCTOR (0.84) and ID (0.80) and shows low PATIENT precision (0.50).

LOCATION is hard for every model: all LLMs reach recall $\leq$\,0.66, with GPT-OSS 120B lowest at 0.58. WEB recall is also modest across the board (uniformly 0.82) despite high precision (0.94--1.00), as shown in the radar profiles of Figure~\ref{fig:llm-benchmark}. These ceiling numbers set expectations for distillation: if the best LLM reaches recall of only 0.66 on LOCATION, the student model is unlikely to surpass this limit.

\subsection{Transformer model comparison on SHIELD}

\begin{figure}[t]
  \centering
  \includegraphics[width=\textwidth,alt={Radar plots of span-level precision and recall for two distilled student models (DeBERTa v3, BioClinical ModernBERT) and two AIMI baselines (v1, v2) evaluated on the SHIELD dataset.}]{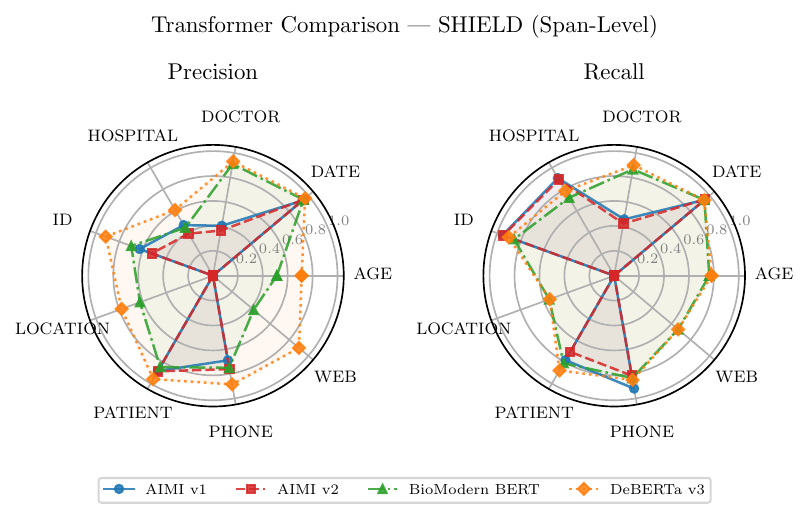}
  \caption{Span-level radar comparison of two distilled student models (DeBERTa v3, BioModern) and two non-student baselines (AIMI v1/v2) on SHIELD. DeBERTa v3 is the best student, achieving the most uniform coverage across all categories. AIMI v1/v2 collapse to zero on AGE, LOCATION, and WEB (absent from their training data), while the distilled students maintain broad coverage.}
  \label{fig:radar-transformer-shield}
\end{figure}

Having established the LLM performance ceiling in Figure~\ref{fig:llm-benchmark}, we compare all four transformer models on the SHIELD gold standard: two distilled students (DeBERTa v3 and BioClinical ModernBERT) and two non-student baselines (AIMI v1 and v2). Figure~\ref{fig:radar-transformer-shield} presents the span-level radar profiles. DeBERTa v3 is the best-performing student model and achieves the most uniform coverage across all \shieldNumCats{} PHI categories. Among the two students, DeBERTa v3 has the higher point estimates on recall for DOCTOR (\debDoctorR{} vs.\ \biomodernDoctorR{}), PATIENT (\debPatientR{} vs.\ \biomodernPatientR{}), and HOSPITAL (\debHospR{} vs.\ \biomodernHospR{}), though none of these differences reach statistical significance after Bonferroni correction. Against the non-student AIMI models, the students hold a large advantage on categories absent from the AIMI training data: AIMI v1 scores 0.00 on AGE, LOCATION, and WEB, and AIMI v2 scores 0.00 on LOCATION and WEB, whereas DeBERTa v3 reaches recall of \debAgeR{}, \debLocR{}, and \debWebR{} on these categories. On shared categories, the AIMI models reach higher HOSPITAL recall (AIMI v1: \aimivOneHospRCI{}) than the SHIELD-trained students (DeBERTa v3: \debHospRCI{}), which marks HOSPITAL as a domain-specific category, while DATE recall is statistically indistinguishable across all models (all within [\dateRecallLo{}--\dateRecallHi{}]). The full forest plot with bootstrap CIs for both precision and recall across all three datasets appears in Figure~\ref{fig:forest-transformers-pr}.

\subsection{Distillation: fidelity and the teacher--student gap}
\label{sec:distillation-results}

\begin{figure}[t]
  \centering
  \includegraphics[width=\textwidth,alt={Dumbbell plot comparing per-category span-level precision and recall between the Gemini 2.5 Flash teacher and the DeBERTa v3 student on SHIELD; green connectors indicate student at-or-above teacher, red indicate teacher advantage.}]{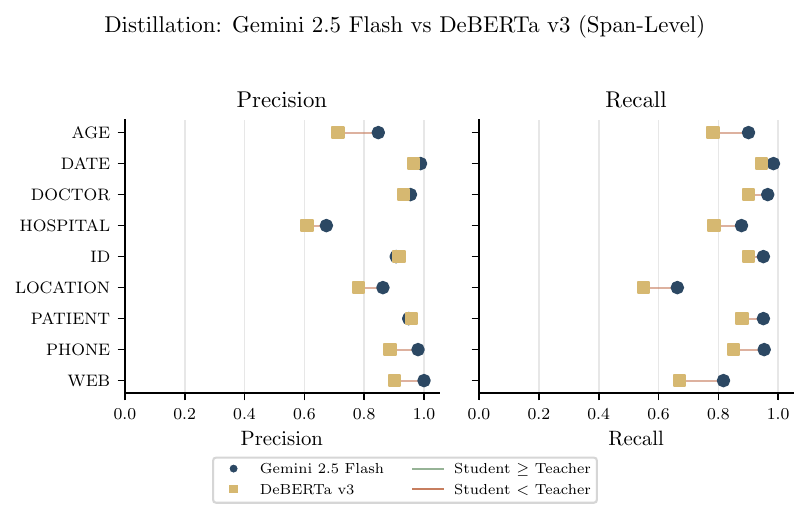}
  \caption{Span-level distillation comparison on SHIELD: Gemini 2.5 Flash (Teacher) vs.\ DeBERTa v3 (Student). Dumbbell plot showing precision (left) and recall (right) per PHI category. Green connectors indicate student $\geq$ teacher; red connectors indicate teacher advantage. Full numerical results with bootstrap 95\% CIs are provided in Table~\ref{tab:distillation} (Appendix~\ref{appendix:supplementary}).}
  \label{fig:dumbbell-distillation}
\end{figure}

Given that DeBERTa v3 was the strongest student in Figure~\ref{fig:radar-transformer-shield}, we examine how well it reproduces its teacher's capabilities. Gemini 2.5 Flash was selected as the distillation teacher on deployment-economics grounds (low cost and high throughput, strong instruction-following, and a large context window) that make large-scale silver labeling practical. Both student architectures were trained on the same \teacherLabelNotesShort{} distilled silver-standard samples; no SHIELD gold-standard annotations were used during training, which preserves the entire SHIELD dataset as an independent held-out evaluation set. Figure~\ref{fig:dumbbell-distillation} presents the teacher-vs.-student comparison for DeBERTa v3. The student trails the teacher on span-level recall across nearly every category, yet retains strong absolute performance: micro-averaged span-level precision \studentMicroP{} and recall \studentMicroR{}, and macro-averaged precision \studentMacroP{} and recall \studentMacroR{}. On a like-for-like macro basis the gap to the teacher is \teacherMacroR{} vs.\ \studentMacroR{} recall (\teacherMacroP{} vs.\ \studentMacroP{} precision). This gap reflects the expected cost of compressing a generative cloud LLM into a \studentParams{}-parameter local model rather than a failure of transfer.

Paired bootstrap significance tests (Bonferroni-corrected, $\alpha = 0.0056$; see Section~\ref{sec:statistical-analysis}) reveal that the recall degradation from teacher to student is significant in eight of the nine PHI categories: AGE ($\distAgeDelta$, $p\distAgeP$), DATE ($\distDateDelta$, $p\distDateP$), DOCTOR ($\distDoctorDelta$, $p\distDoctorP$), HOSPITAL ($\distHospitalDelta$, $p\distHospitalP$), ID ($\distIdDelta$, $p\distIdP$), LOCATION ($\distLocationDelta$, $p\distLocationP$), PATIENT ($\distPatientDelta$, $p\distPatientP$), and PHONE ($\distPhoneDelta$, $p\distPhoneP$). Only WEB ($\distWebDelta$, $p\distWebP$) fails to reach significance after correction, owing to its small support (\shieldWebSpans{} gold spans) and wide bootstrap CI (student recall \debWebR{} [0.52--0.83]). No category shows a significant improvement. The uniformly negative deltas indicate that the student trails its generative teacher on recall across the board, consistent with the cost of compressing contextual reasoning into a token-classification model. The student still reaches strong absolute recall on high-support structured categories (DATE \debDateR{}, DOCTOR \debDoctorR{}, PATIENT \debPatientR{}), and its remaining weak spot, LOCATION (\debLocR{} [0.49--0.61], \shieldLocationSpans{} gold spans), sits close to the LLM ceiling itself ($\leq$\,0.66 for every benchmarked LLM).

\subsection{Cross-dataset generalization}

\begin{figure}[!ht]
  \centering
  \includegraphics[width=\textwidth,height=0.82\textheight,keepaspectratio,alt={Forest plot of span-level precision and recall with bootstrap 95 percent confidence intervals across SHIELD, i2b2, and AIMI for four transformer models (DeBERTa v3, BioClinical ModernBERT, AIMI v1, AIMI v2).}]{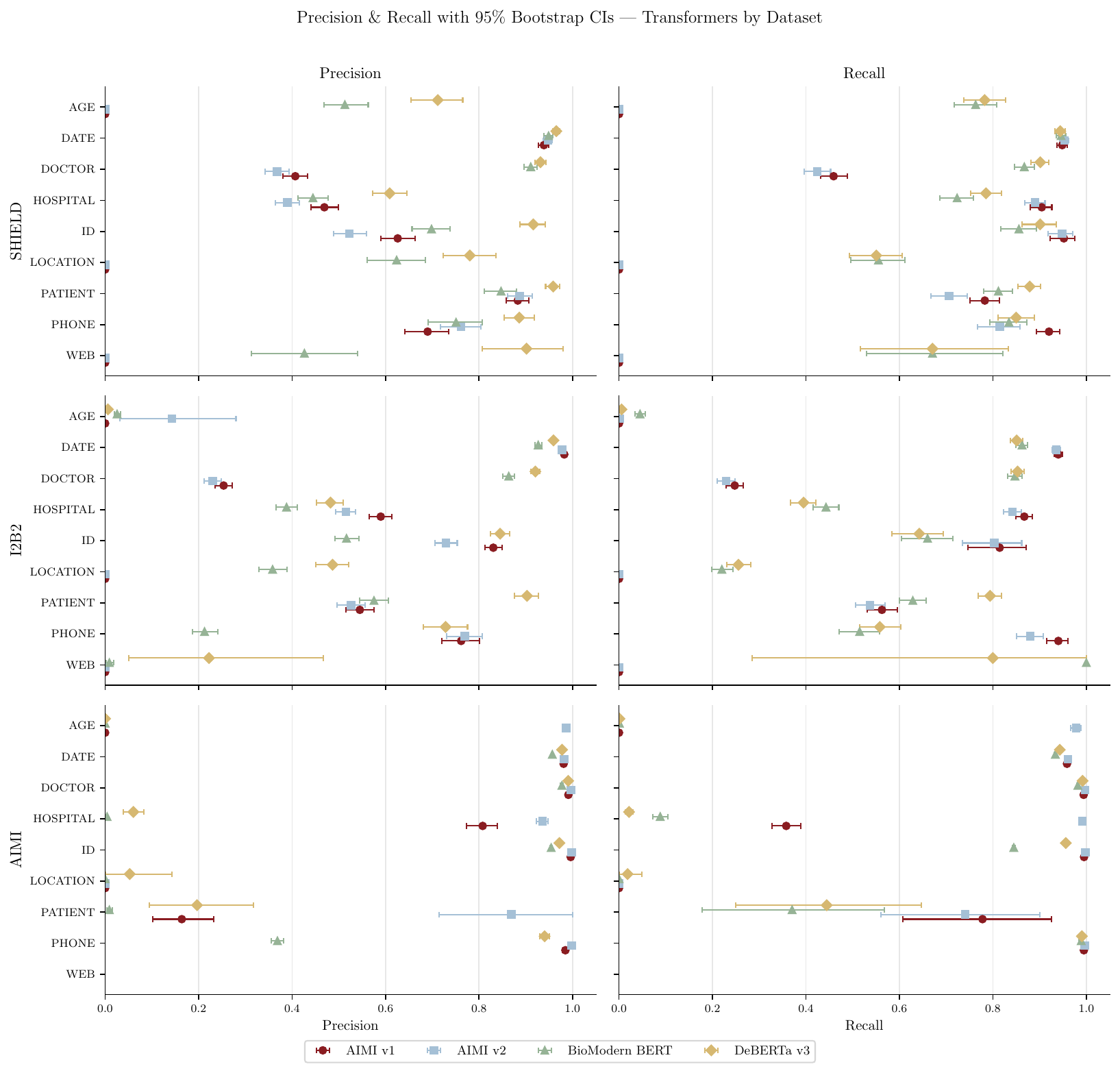}
  \caption{Span-level precision (left) and recall (right) with bootstrap 95\% CIs for four transformer models across all three evaluation datasets (SHIELD, i2b2, AIMI). DeBERTa v3 achieves the most balanced coverage on SHIELD, while AIMI v1/v2 show zero recall on categories absent from their training data. Precision patterns reveal complementary trade-offs: AIMI models achieve high precision on their trained categories but produce no predictions (and thus undefined or zero precision) on unseen categories, whereas SHIELD-trained models maintain moderate precision across all categories. Full numerical results are in Tables~\ref{tab:transformer-shield},~\ref{tab:cross-i2b2}, and~\ref{tab:cross-aimi} (Appendix~\ref{appendix:supplementary}).}
  \label{fig:forest-transformers-pr}
\end{figure}

To assess whether SHIELD-trained models generalize beyond their training domain, we evaluated all four transformer models on i2b2 2014 and AIMI at span level. The forest plot in Figure~\ref{fig:forest-transformers-pr} visualizes precision and recall with bootstrap 95\% CIs across all three datasets simultaneously; full per-category numerical results are provided in Tables~\ref{tab:cross-i2b2} and~\ref{tab:cross-aimi} (Appendix~\ref{appendix:supplementary}), along with per-dataset radar profiles (Figures~\ref{fig:radar-transformer-i2b2} and~\ref{fig:radar-transformer-aimi}).

\textbf{Generalization to i2b2.} DeBERTa v3 leads on person-name categories: DOCTOR (0.85 [0.84--0.87]) and PATIENT (0.79 [0.77--0.82]), which shows that SHIELD-trained models generalize to i2b2's clinical notes. AIMI v1/v2 reach strong DATE recall (0.94 [0.93--0.95]) since dates are universal, but do not detect LOCATION (0.00), which was absent from their training data. All models score near zero on AGE due to label-definition differences between SHIELD and i2b2. HOSPITAL shows an asymmetry: AIMI models reach high recall (0.87 [0.85--0.88] for AIMI v1) with non-overlapping CIs against SHIELD-trained models (DeBERTa v3: \debHospItwobR{} [0.37--0.42]), which indicates that HOSPITAL recognition does not transfer across domains. These results are consistent with cross-institute generalization findings \citep{yang_study_2019}.

\textbf{Generalization to AIMI.} On AIMI's home dataset, AIMI v2 reaches near-perfect recall on its core categories: DOCTOR 1.00 [1.00--1.00], ID 1.00 [1.00--1.00], and HOSPITAL 0.99 [0.99--1.00]. Only AIMI v2 detects AGE on AIMI (0.98 [0.97--0.99]); all other models score 0.00. DeBERTa v3 stays competitive on universal categories, DATE (0.94 [0.94--0.95]), ID (0.96 [0.95--0.96]), and PHONE (0.99 [0.99--0.99]), but shows limited transfer on institution-specific HOSPITAL (0.02 [0.01--0.03]). Low-support categories produce wide intervals: PATIENT has only 27 spans, yielding CIs spanning 30+ percentage points (e.g., AIMI v1: 0.78 [0.61--0.93]). LOCATION is near-zero across all models.

Together, the cross-dataset results in Figure~\ref{fig:forest-transformers-pr} show an asymmetric generalization pattern: SHIELD-trained models (DeBERTa v3, BioModern) generalize well to shared structured categories across both datasets, while domain-specific models (AIMI v1/v2) excel on their home data but do not generalize to categories absent from their training. Institution-specific entities (HOSPITAL, LOCATION) are the hardest to transfer in both directions, whereas universal structured categories (DATE, ID, PHONE) transfer reliably regardless of training data. This conclusion is strongest in the clean i2b2 direction, where the label taxonomy matches. The AIMI HOSPITAL and LOCATION comparisons are partly confounded by label remapping (AIMI's \texttt{VENDOR}\,$\to$\,unified \texttt{HOSPITAL} and AIMI's \texttt{HOSPITAL}\,$\to$\,unified \texttt{LOCATION}; Table~\ref{tab:mapping-aimi}), so those two categories compare semantically different entity types and should be read with that caveat.

\clearpage

\section{Discussion}

The results support two key observations for clinical informatics: the value of semantic diversity in benchmarking, and the viability of SLM distillation for secure, enterprise-scale de-identification.

\textbf{Design rationale.} Our pipeline uses two deliberate selection stages rather than post-hoc ablations. First, the four-way LLM comparison (Figure~\ref{fig:llm-benchmark}) establishes the current-LLM performance ceiling on SHIELD; the distillation teacher (Gemini 2.5 Flash) is both competitive at this ceiling and chosen on deployment-economics grounds (cost, throughput, instruction-following, and context window), since at-scale silver labeling is governed by inference economics more than by marginal accuracy differences. Second, the choice of student architectures (DeBERTa v3 and BioClinical ModernBERT) was informed by \citet{antoun_modernbert_2025}, whose controlled comparison showed DeBERTa v3 offers better sample efficiency while ModernBERT offers long-context and inference-speed advantages. Further ablation on distillation data volume and multi-teacher ensembling are directions for future work.

\textbf{Distillation effectiveness.} Paired bootstrap tests (Table~\ref{tab:distillation}, Appendix~\ref{appendix:supplementary}) show that the student's recall is significantly below the teacher's in eight of nine PHI categories after Bonferroni correction; only WEB ($\distWebDelta$, $p\distWebP$), with its small support and wide bootstrap CI, fails to reach significance. No category shows a significant improvement. This uniform recall gap is consistent with the cross-institute generalization findings of \citet{yang_study_2019}, who observed similar degradation when transferring de-identification capability across settings. The student retains strong absolute performance, micro-averaged span-level precision of \studentMicroP{} and recall of \studentMicroR{} (macro-averaged \studentMacroP{}/\studentMacroR{}), and on HOSPITAL the teacher and student precision CIs overlap (teacher: 0.67 [0.64--0.71] vs.\ student: 0.61 [0.57--0.65]). The systematic but bounded recall loss is the expected cost of compressing a generative teacher into a token-classification student; targeted error correction \citep{lee_targeted_2025} is a direction for closing the remaining gap.

\textbf{Cost advantage of distillation.} Our evaluation of four LLMs (Figure~\ref{fig:llm-benchmark}) shows that all perform well on SHIELD, but none solves the problem universally: LOCATION recall remains $\leq$\,0.66 for every LLM, and WEB recall is uniformly 0.82. The open-weight GPT-OSS models perform on par with the proprietary Gemini models, which indicates that local LLM deployment is technically feasible, but even the smaller GPT-OSS 20B requires substantial GPU infrastructure. The distillation approach avoids this trade-off. The STARR-OMOP warehouse contains approximately \starrChars{} characters ($\sim$\starrTokens{} tokens at 4 characters per token) of clinical text. Processing this entire volume through Gemini 2.5 Flash on Vertex AI at Flex/Batch pricing \citep{google_vertexai_pricing_2025} would cost an estimated \$23.7k in input tokens (\$0.15 per 1M tokens $\times$ \starrTokensShort{} tokens) plus \$670k in output tokens (\$1.25 per 1M tokens $\times$ 536B estimated output tokens), totaling approximately \$694k (see Appendix~\ref{appendix:cost-analysis} for the full derivation). The output token estimate is a conservative lower bound derived from scaling the structured JSON output of the AIMI v2 de-identification pipeline to the full warehouse; actual costs could be higher depending on output verbosity. By contrast, our distillation approach queries Gemini 2.5 Flash only once, on \teacherLabelNotesShort{} notes, after which the DeBERTa v3 model runs locally at negligible marginal cost and is not re-billed per note. The GPT-OSS models, being open-weight, can run on-premise, but at 20--120B parameters they still require significant GPU compute, far more than the \studentParams{}-parameter DeBERTa v3. Recent work on privacy-preserving local LLMs \citep{wiest_deidentifying_2025} and LLM-powered de-identification frameworks \citep{singh_redactor_2025} points to growing interest in cost-effective alternatives to cloud-dependent pipelines. At enterprise scale, efforts to de-identify billions of notes \citep{kocaman_automated_2025} and optimize institutional pipelines \citep{prakash_improving_2025} add further support for efficient local models.

\textbf{Domain specificity and benchmark diversity.} The AIMI v1/v2 models fail to detect WEB and LOCATION (0.00 across all metrics on SHIELD) and reach near-zero AGE recall, which shows the importance of training data breadth. Cross-dataset evaluation (Tables~\ref{tab:cross-i2b2} and~\ref{tab:cross-aimi}, Appendix~\ref{appendix:supplementary}) quantifies this asymmetric generalization: SHIELD-trained DeBERTa v3 stays competitive (precision/recall) on universal categories across both i2b2 (DOCTOR: 0.92/0.85, PATIENT: 0.90/0.79) and AIMI (DATE: 0.98/0.94, ID: 0.97/0.96), as shown in Figure~\ref{fig:forest-transformers-pr}, while AIMI models show limited generalization on out-of-distribution categories (LOCATION: 0.00 on i2b2). Institution-specific entities are hardest to transfer: HOSPITAL recall for DeBERTa v3 drops to \debHospItwobR{} on i2b2 and \debHospAimiR{} on AIMI, which shows that hospital-name patterns are highly domain-dependent. The AIMI cross-dataset figures for HOSPITAL and LOCATION should be read with the label mapping caveat in mind (Table~\ref{tab:mapping-aimi}): AIMI's \texttt{VENDOR} entities map to unified \texttt{HOSPITAL} and AIMI's \texttt{HOSPITAL} entities map to unified \texttt{LOCATION}, so these cross-dataset comparisons reflect semantically different entity types, which partly explains the low transfer on both categories. Surrogate replacement strategies such as Markov-chain-based approaches \citep{osborne_markov_2025} offer complementary methods for preserving text utility after de-identification. SHIELD's distinctiveness reflects genuine differences in clinical content: its diversity sampling captures procedural, nursing, and patient-facing documentation absent from legacy discharge-summary-dominated corpora, which requires de-identification models to handle PHI patterns in operative reports, care plans, and patient instructions.

\textbf{The case for note-type-specialized models.} While diverse training data improves coverage across PHI categories, the strong performance of AIMI v1/v2 on their home domain (Table~\ref{tab:cross-aimi}, Appendix~\ref{appendix:supplementary}; Figure~\ref{fig:forest-transformers-pr}: AIMI v2 precision/recall of DOCTOR 1.00/1.00, ID 1.00/1.00, HOSPITAL 0.94/0.99) suggests that for high-volume, semi-structured note types such as radiology and pathology reports, specialized models trained on domain-specific data may outperform general-purpose alternatives. These note types follow predictable templates with consistent PHI placement, which a narrowly trained model can exploit. A practical deployment strategy may therefore combine a broad-coverage model like DeBERTa v3 for heterogeneous clinical notes with specialized models for major semi-structured note types, routing documents by note type to the appropriate model. This hybrid approach draws on the strengths of both: diversity-trained models for robustness across the long tail of clinical text, and domain-specific models for accuracy on the most common structured formats.

\textbf{Scope.} The primary contributions of this work are the release of the SHIELD dataset and the distilled DeBERTa v3 model, together with a distillation framework for producing locally deployable de-identification SLMs. Our evaluation therefore focuses on characterizing LLM teacher performance and the fidelity of teacher-to-student knowledge transfer, rather than benchmarking existing de-identification systems end to end. A systematic comparison of established pipelines and software (rule-based systems such as Philter, prior neural architectures such as NeuroNER, and commercial APIs) is a direction best addressed in a dedicated benchmarking study using SHIELD and other datasets. Instructions for accessing both the dataset and the model are provided at \url{https://github.com/susom/shield_dataset}, and we aim to enable such comparative evaluations by the broader community.

\textbf{Limitations.} This work has three main limitations. First, SHIELD contains only \shieldNotes{} notes, which limits statistical power for rare PHI categories. Bootstrap 95\% CIs quantify this uncertainty: WEB recall for DeBERTa v3 on SHIELD is 0.67 [0.52--0.83], a 30-percentage-point range reflecting only \shieldWebSpans{} gold-standard spans, and LOCATION recall is 0.55 [0.49--0.61] (\shieldLocationSpans{} spans). High-support categories yield tight CIs: DATE recall is 0.94 [0.93--0.95] (\shieldDateSpans{} spans). On AIMI, PATIENT (27 spans) produces CIs spanning 30+ percentage points, which makes model comparisons on this category unreliable. Second, all notes originate from a single institution (Stanford Health Care via STARR-OMOP), and multi-site validation is needed to confirm generalizability claims. Third, our distillation uses a single teacher model; ensembling multiple teachers or adding targeted error correction \citep{lee_targeted_2025} could improve student accuracy on hard categories.

\section{Conclusion}

The SHIELD dataset provides a modern standard for clinical de-identification benchmarks, prioritizing demographic and document-type diversity within a single health system over sheer size. Paired with our teacher-student distillation framework, it shows that institutions need not rely on expensive, privacy-restricted cloud APIs or large local GPU clusters. Our distilled DeBERTa v3 model reaches micro-averaged span-level precision of \studentMicroP{} and recall of \studentMicroR{} on SHIELD (macro-averaged recall \studentMacroR{} vs.\ the teacher's \teacherMacroR{}). It trails its cloud-based Gemini 2.5 Flash teacher on per-category recall but runs entirely on local enterprise hardware at a fraction of the cost. Cross-dataset evaluation on i2b2 and AIMI shows that diversity-trained models generalize well on universal, structured PHI categories (DATE, ID, PHONE), while institution-specific entities (HOSPITAL, LOCATION) remain resistant to cross-domain transfer regardless of training diversity. This pattern is cleanest in the i2b2 direction, since the AIMI HOSPITAL/LOCATION comparison is confounded by label remapping (Table~\ref{tab:mapping-aimi}). Domain-specific models such as AIMI v1/v2 reach near-perfect accuracy on their home data but do not generalize to PHI types absent from their training. These findings suggest that optimal enterprise deployment may benefit from a hybrid strategy that pairs broad-coverage models for heterogeneous clinical text with specialized models for high-volume, semi-structured note types. Access instructions for the SHIELD dataset and the distilled DeBERTa v3 model are provided at \url{https://github.com/susom/shield_dataset} to lower the barrier to safe, scalable clinical text de-identification.

\begin{ack}
This research was funded, in part, by the Advanced Research Projects Agency for Health (ARPA-H) under award number AY2AX000024. The views and conclusions contained in this document are those of the authors and should not be interpreted as representing the official policies, either expressed or implied, of the U.S. Government.

We gratefully acknowledge Professor Sylvia Plevritis, Ph.D. (Stanford University), Principal Investigator of this project, for her support of this work. We also thank Todd Ferris (Deputy Chief Information Officer, Stanford University School of Medicine) and Anthea Buchin (Director of Research Technology Application Solutions, Stanford Healthcare) for their leadership and institutional support of this work. We also thank our colleagues Joe Pallas, Hannah Morgan-Cooper, Farnoosh Sheikhi, Shikha Kothari, and Alvaro Alvarez, all of Stanford Healthcare, for their contributions to the research data infrastructure and engineering work that enabled this project.
\end{ack}

\bibliographystyle{plainnat}
\bibliography{references}

\begin{thebibliography}{30}
\providecommand{\natexlab}[1]{#1}
\providecommand{\url}[1]{\texttt{#1}}
\expandafter\ifx\csname urlstyle\endcsname\relax
  \providecommand{\doi}[1]{doi: #1}\else
  \providecommand{\doi}{doi: \begingroup \urlstyle{rm}\Url}\fi

\bibitem[Antoun et~al.(2025)Antoun, Sagot, and Seddah]{antoun_modernbert_2025}
Wissam Antoun, Beno{\^i}t Sagot, and Djam{\'e} Seddah.
\newblock {ModernBERT} or {DeBERTaV3}? {Examining} architecture and data
  influence on transformer encoder models performance.
\newblock In \emph{Proceedings of the 14th International Joint Conference on
  Natural Language Processing and the 4th Conference of the Asia-Pacific
  Chapter of the Association for Computational Linguistics (IJCNLP-AACL 2025)},
  2025.
\newblock URL \url{https://aclanthology.org/2025.ijcnlp-long.164/}.

\bibitem[Belcak et~al.(2025)Belcak, Heinrich, Diao, Fu, Dong, Muralidharan,
  Lin, and Molchanov]{belcak_small_2025}
Peter Belcak, Greg Heinrich, Shizhe Diao, Yonggan Fu, Xin Dong, Saurav
  Muralidharan, Yingyan~Celine Lin, and Pavlo Molchanov.
\newblock Small {Language} {Models} are the {Future} of {Agentic} {AI}, June
  2025.
\newblock URL \url{http://arxiv.org/abs/2506.02153}.
\newblock arXiv:2506.02153 [cs].

\bibitem[Callahan et~al.(2023)Callahan, Ashley, Datta, Desai, Ferris, Fries,
  Halaas, Langlotz, Mackey, Posada, Pfeffer, and Shah]{callahan2023starr}
Alison Callahan, Euan Ashley, Somalee Datta, Priyamvada Desai, Todd~A Ferris,
  Jason~A Fries, Michael Halaas, Curtis~P Langlotz, Sean Mackey, José~D
  Posada, Michael~A Pfeffer, and Nigam~H Shah.
\newblock The stanford medicine data science ecosystem for clinical and
  translational research.
\newblock \emph{JAMIA Open}, 6\penalty0 (3):\penalty0 ooad054, 08 2023.
\newblock ISSN 2574-2531.
\newblock \doi{10.1093/jamiaopen/ooad054}.
\newblock URL \url{https://doi.org/10.1093/jamiaopen/ooad054}.

\bibitem[Chambon et~al.(2023)Chambon, Wu, Steinkamp, Adleberg, Cook, and
  Langlotz]{chambon_automated_2023}
Pierre~J Chambon, Christopher Wu, Jackson~M Steinkamp, Jason Adleberg, Tessa~S
  Cook, and Curtis~P Langlotz.
\newblock Automated deidentification of radiology reports combining transformer
  and ``hide in plain sight'' rule-based methods.
\newblock \emph{Journal of the American Medical Informatics Association},
  30\penalty0 (2):\penalty0 318--328, January 2023.
\newblock ISSN 1067-5027, 1527-974X.
\newblock \doi{10.1093/jamia/ocac219}.
\newblock URL \url{https://academic.oup.com/jamia/article/30/2/318/6843283}.

\bibitem[Datta et~al.(2020)Datta, Posada, Olson, Li, O'Reilly, Balraj,
  Mesterhazy, Pallas, Desai, and
  Shah]{datta2020newparadigmacceleratingclinical}
Somalee Datta, Jose Posada, Garrick Olson, Wencheng Li, Ciaran O'Reilly, Deepa
  Balraj, Joseph Mesterhazy, Joseph Pallas, Priyamvada Desai, and Nigam Shah.
\newblock A new paradigm for accelerating clinical data science at stanford
  medicine, 2020.
\newblock URL \url{https://arxiv.org/abs/2003.10534}.

\bibitem[Eyre et~al.(2025)Eyre, Gan, Hu, Bowles, Stanley, Shi, DuVall, and
  Alba]{eyre_evaluating_2025}
Hannah Eyre, Qiwei Gan, Mengke Hu, Annie Bowles, Johnathan Stanley, Jianlin
  Shi, Scott~L. DuVall, and Patrick~R. Alba.
\newblock Evaluating {Clinical} {Note} {Deidentification} {Tools} and
  {Transformer} {Transferability} between {Public} and {Private} {Data} from
  the {US} {Department} of {Veterans} {Affairs}, June 2025.
\newblock URL
  \url{https://www.medrxiv.org/content/10.1101/2025.03.21.25323520v2}.

\bibitem[{Google Cloud}(2025)]{google_vertexai_pricing_2025}
{Google Cloud}.
\newblock Vertex {AI} — {Generative} {AI} {Pricing}.
\newblock \url{https://cloud.google.com/vertex-ai/generative-ai/pricing}, 2025.
\newblock Gemini 2.5 Flash Flex/Batch pricing: \$0.15 per 1M input tokens,
  \$1.25 per 1M output tokens. Accessed March 2025.

\bibitem[He et~al.(2021)He, Gao, and Chen]{he2021deberta}
Pengcheng He, Jianfeng Gao, and Weizhu Chen.
\newblock Debertav3: Improving deberta using electra-style pre-training with
  gradient-disentangled embedding sharing.
\newblock arXiv preprint arXiv:2111.09543, 2021.

\bibitem[Heider and Meystre(2024)]{heider_extensible_2024-2}
Paul~M Heider and St{\'e}phane~M Meystre.
\newblock An {Extensible} {Evaluation} {Framework} {Applied} to {Clinical}
  {Text} {Deidentification} {Natural} {Language} {Processing} {Tools}:
  {Multisystem} and {Multicorpus} {Study}.
\newblock \emph{Journal of Medical Internet Research}, 26:\penalty0 e55676, May
  2024.
\newblock ISSN 1439-4456.
\newblock \doi{10.2196/55676}.
\newblock URL \url{https://pmc.ncbi.nlm.nih.gov/articles/PMC11167315/}.

\bibitem[Kansal et~al.(2025)Kansal, Chen, Jin, Rajpurkar, and
  Kim]{kansal2025mcmed}
Aman Kansal, Emma Chen, Boyang~Tom Jin, Pranav Rajpurkar, and David~A. Kim.
\newblock {MC-MED}, multimodal clinical monitoring in the emergency department.
\newblock \emph{Scientific Data}, 12\penalty0 (1):\penalty0 1094, 2025.
\newblock \doi{10.1038/s41597-025-05419-5}.
\newblock URL \url{https://doi.org/10.1038/s41597-025-05419-5}.

\bibitem[Kather et~al.(2024)Kather, Ferber, Wiest, Gilbert, and
  Truhn]{kather2024llm}
Jakob~Nikolas Kather, Dyke Ferber, Isabelle~C. Wiest, Stephen Gilbert, and
  Daniel Truhn.
\newblock Large language models could make natural language again the universal
  interface of healthcare.
\newblock \emph{Nature Medicine}, 30:\penalty0 2708--2710, 2024.
\newblock \doi{10.1038/s41591-024-03259-5}.

\bibitem[Kocaman et~al.(2023)Kocaman, Haq, and Talby]{kocaman_beyond_2023}
Veysel Kocaman, Hasham~Ul Haq, and David Talby.
\newblock Beyond {Accuracy}: {Automated} {De}-{Identification} of {Large}
  {Real}-{World} {Clinical} {Text} {Datasets}, December 2023.
\newblock URL \url{http://arxiv.org/abs/2312.08495}.
\newblock arXiv:2312.08495 [cs].

\bibitem[Kocaman et~al.(2025)Kocaman, Mico, Kaya, Taiyab, Talby, Surh, Guo,
  Tomer, and Kramer]{kocaman_automated_2025}
Veysel Kocaman, Lindsay Mico, Mustafa~Aytug Kaya, Nadaa Taiyab, David Talby,
  Tae Surh, Yuqing Guo, Vivek Tomer, and Robert Kramer.
\newblock Automated {De}-{Identification}, {Consistent} {Obfuscation}, and
  {Regulatory} {Grade} {Validation} of 2 {Billion} {Patient} {Notes}, September
  2025.
\newblock URL \url{https://www.researchsquare.com/article/rs-6867162/v1}.

\bibitem[Kovačević et~al.(2024)Kovačević, Bašaragin, Milošević, and
  Nenadić]{kovacevic_-identification_2024}
Aleksandar Kovačević, Bojana Bašaragin, Nikola Milošević, and Goran
  Nenadić.
\newblock De-identification of clinical free text using natural language
  processing: {A} systematic review of current approaches.
\newblock \emph{Artificial Intelligence in Medicine}, 151:\penalty0 102845, May
  2024.
\newblock ISSN 09333657.
\newblock \doi{10.1016/j.artmed.2024.102845}.
\newblock URL
  \url{https://linkinghub.elsevier.com/retrieve/pii/S0933365724000873}.

\bibitem[Kraljevic et~al.(2023)Kraljevic, Shek, Yeung, Sheldon, Al-Agil,
  Shuaib, Bai, Noor, Shah, Dobson, and Teo]{kraljevic_validating_2023}
Zeljko Kraljevic, Anthony Shek, Joshua~Au Yeung, Ewart~Jonathan Sheldon,
  Mohammad Al-Agil, Haris Shuaib, Xi~Bai, Kawsar Noor, Anoop~D. Shah, Richard
  Dobson, and James Teo.
\newblock Validating transformers for redaction of text from electronic health
  records in real-world healthcare, October 2023.
\newblock URL \url{http://arxiv.org/abs/2310.04468}.
\newblock arXiv:2310.04468 [cs].

\bibitem[Kuo et~al.(2025)Kuo, Soltan, O'Hanlon, Hasanic, Clifton, Gary,
  Furniss, and Eyre]{kuo_benchmarking_2025}
Rachel Kuo, Andrew A.~S. Soltan, Ciaran O'Hanlon, Alan Hasanic, David~A.
  Clifton, Collins Gary, Dominic Furniss, and David~W. Eyre.
\newblock Benchmarking transformer-based models for medical record
  deidentification: {A} single centre, multi-specialty evaluation, May 2025.
\newblock URL
  \url{https://www.medrxiv.org/content/10.1101/2025.05.05.25326979v1}.

\bibitem[Lee et~al.(2025)Lee, Guo, Jin, and Goudarzi]{lee_targeted_2025}
Hee-Jin Lee, Zhen Guo, Luchao Jin, and Morteza~Moazami Goudarzi.
\newblock Targeted {Error} {Correction} in {Knowledge} {Distillation}: {Small}
  {Language} {Models} {Surpass} {GPT}, November 2025.
\newblock URL \url{http://arxiv.org/abs/2511.03005}.
\newblock arXiv:2511.03005 [cs].

\bibitem[Liu et~al.(2023)Liu, Huang, Yu, Zhang, Wu, Cao, Dai, Zhao, Li, Shu,
  Zeng, Sun, Liu, Shen, Li, Liu, Zhu, and Li]{liu_deid-gpt_2023}
Zhengliang Liu, Yue Huang, Xiaowei Yu, Lu~Zhang, Zihao Wu, Chao Cao, Haixing
  Dai, Lin Zhao, Yiwei Li, Peng Shu, Fang Zeng, Lichao Sun, Wei Liu, Dinggang
  Shen, Quanzheng Li, Tianming Liu, Dajiang Zhu, and Xiang Li.
\newblock {DeID}-{GPT}: {Zero}-shot {Medical} {Text} {De}-{Identification} by
  {GPT}-4, December 2023.
\newblock URL \url{http://arxiv.org/abs/2303.11032}.
\newblock arXiv:2303.11032 [cs].

\bibitem[{Meta AI}(2026)]{llama4herd_2026}
{Meta AI}.
\newblock The {Llama} 4 {Herd}: {Architecture}, {Training}, {Evaluation}, and
  {Deployment} {Notes}.
\newblock arXiv preprint arXiv:2601.11659, 2026.
\newblock URL \url{https://arxiv.org/abs/2601.11659}.
\newblock Mixture-of-Experts; released variants Scout and Maverick. Llama 4
  Community License.

\bibitem[Negash et~al.(2023)Negash, Katz, Neilson, Moni, Nesca, Singer, and
  Enns]{negash_-identification_2023}
Bekelu Negash, Alan Katz, Christine~J. Neilson, Moniruzzaman Moni, Marcello
  Nesca, Alexander Singer, and Jennifer~E. Enns.
\newblock De-identification of {Free} {Text} {Data} containing {Personal}
  {Health} {Information}: {A} {Scoping} {Review} of {Reviews}.
\newblock \emph{International Journal of Population Data Science}, 8\penalty0
  (1), December 2023.
\newblock ISSN 2399-4908.
\newblock \doi{10.23889/ijpds.v8i1.2153}.
\newblock URL \url{https://ijpds.org/article/view/2153}.

\bibitem[{OpenAI}(2025)]{openai_gptoss_2025}
{OpenAI}.
\newblock {GPT-OSS}: {Open}-weight models for reasoning, agentic tasks, and
  versatile developer use cases.
\newblock arXiv preprint arXiv:2508.10925, 2025.
\newblock URL \url{https://huggingface.co/openai/gpt-oss-120b}.
\newblock Mixture-of-Experts architecture; GPT-OSS 120B (117B total, 5.1B
  active parameters) and GPT-OSS 20B (21B total, 3.6B active parameters).
  Apache 2.0 license.

\bibitem[Osborne et~al.(2025)Osborne, Trotter, O'Leary, Coffee, Cochran,
  Mansilla-Gonzalez, Nadimpalli, McAnnally, Almudaifer, Curtis, Aly, and
  Kennedy]{osborne_markov_2025}
John~D. Osborne, Andrew Trotter, Tobias O'Leary, Chris Coffee, Micah~D.
  Cochran, Luis Mansilla-Gonzalez, Akhil Nadimpalli, Alex McAnnally,
  Abdulateef~I. Almudaifer, Jeffrey~R. Curtis, Salma~M. Aly, and Richard~E.
  Kennedy.
\newblock A {Markov} {Chain} {Replacement} {Strategy} for {Surrogate}
  {Identifiers}: {Minimizing} {Re}-{Identification} {Risk} {While} {Preserving}
  {Text} {Reuse}.
\newblock \emph{Electronics}, 14\penalty0 (19):\penalty0 3945, October 2025.
\newblock ISSN 2079-9292.
\newblock \doi{10.3390/electronics14193945}.
\newblock URL \url{https://pmc.ncbi.nlm.nih.gov/articles/PMC12536513/}.

\bibitem[Prakash et~al.(2025)Prakash, Attias, Chambon, Xu, Truong, Delbrouck,
  Cook, and Langlotz]{prakash_improving_2025}
Eva Prakash, Maayane Attias, Pierre Chambon, Justin Xu, Steven Truong,
  Jean-Benoit Delbrouck, Tessa Cook, and Curtis Langlotz.
\newblock Improving the {Performance} of {Radiology} {Report}
  {De}-identification with {Large}-{Scale} {Training} and {Benchmarking}
  {Against} {Cloud} {Vendor} {Methods}, November 2025.
\newblock URL \url{http://arxiv.org/abs/2511.04079}.
\newblock arXiv:2511.04079 [cs].

\bibitem[Singh et~al.(2025)Singh, Dzialo, Kim, Srivatsa, Bulu, Gadde, and
  Kenthapadi]{singh_redactor_2025}
Praphul Singh, Charlotte Dzialo, Jangwon Kim, Sumana Srivatsa, Irfan Bulu, Sri
  Gadde, and Krishnaram Kenthapadi.
\newblock {RedactOR}: {An} {LLM}-{Powered} {Framework} for {Automatic}
  {Clinical} {Data} {De}-{Identification}, July 2025.
\newblock URL \url{http://arxiv.org/abs/2505.18380}.
\newblock arXiv:2505.18380 [cs].

\bibitem[Sounack et~al.(2025)Sounack, Davis, Durieux, Chaffin, Pollard, Lehman,
  Johnson, McDermott, Naumann, and Lindvall]{sounack_bioclinical_2025}
Thomas Sounack, Joshua Davis, Brigitte Durieux, Antoine Chaffin, Tom~J.
  Pollard, Eric Lehman, Alistair E.~W. Johnson, Matthew McDermott, Tristan
  Naumann, and Charlotta Lindvall.
\newblock {BioClinical} {ModernBERT}: {A} {State}-of-the-{Art} {Long}-{Context}
  {Encoder} for {Biomedical} and {Clinical} {NLP}, June 2025.
\newblock URL \url{http://arxiv.org/abs/2506.10896}.
\newblock arXiv:2506.10896 [cs].

\bibitem[Stubbs and Uzuner(2015)]{stubbs2015corpus}
Amber Stubbs and {"O}zlem Uzuner.
\newblock Annotating longitudinal clinical narratives for de-identification:
  The 2014 i2b2/uthealth corpus.
\newblock \emph{Journal of Biomedical Informatics}, 58:\penalty0 S20--S29,
  2015.
\newblock \doi{10.1016/j.jbi.2015.07.020}.
\newblock PMID: 26319540.

\bibitem[Uzuner et~al.(2007)Uzuner, Luo, and Szolovits]{uzuner_evaluating_2007}
{\"O}zlem Uzuner, Yuan Luo, and Peter Szolovits.
\newblock Evaluating the {State}-of-the-{Art} in {Automatic}
  {De}-identification.
\newblock \emph{Journal of the American Medical Informatics Association},
  14\penalty0 (5):\penalty0 550--563, September 2007.
\newblock ISSN 1067-5027.
\newblock \doi{10.1197/jamia.M2444}.
\newblock URL \url{https://doi.org/10.1197/jamia.M2444}.

\bibitem[Wiest et~al.(2025)Wiest, Le{\ss}mann, Wolf, Ferber, Treeck, Zhu,
  Ebert, Westphalen, Wermke, and Kather]{wiest_deidentifying_2025}
Isabella~C. Wiest, Marie-Elisabeth Le{\ss}mann, Fabian Wolf, Dyke Ferber,
  Marko~Van Treeck, Jiefu Zhu, Matthias~P. Ebert, Christoph~Benedikt
  Westphalen, Martin Wermke, and Jakob~Nikolas Kather.
\newblock Deidentifying {Medical} {Documents} with {Local},
  {Privacy}-{Preserving} {Large} {Language} {Models}: {The} {LLM}-{Anonymizer}.
\newblock \emph{NEJM AI}, 2\penalty0 (4):\penalty0 AIdbp2400537, March 2025.
\newblock \doi{10.1056/AIdbp2400537}.
\newblock URL \url{https://ai.nejm.org/doi/full/10.1056/AIdbp2400537}.

\bibitem[Williamson and Shmoys(2011)]{williamson2011design}
David~P Williamson and David~B Shmoys.
\newblock \emph{The design of approximation algorithms}.
\newblock Cambridge university press, 2011.

\bibitem[Yang et~al.(2019)Yang, Lyu, Li, Lee, Bian, Hogan, and
  Wu]{yang_study_2019}
Xi~Yang, Tianchen Lyu, Qian Li, Chih-Yin Lee, Jiang Bian, William~R. Hogan, and
  Yonghui Wu.
\newblock A study of deep learning methods for de-identification of clinical
  notes in cross-institute settings.
\newblock \emph{BMC Medical Informatics and Decision Making}, 19\penalty0
  (Suppl 5):\penalty0 232, December 2019.
\newblock ISSN 1472-6947.
\newblock \doi{10.1186/s12911-019-0935-4}.
\newblock URL \url{https://pmc.ncbi.nlm.nih.gov/articles/PMC6894104/}.

\end{thebibliography}

\appendix

\section{Supplementary Tables and Figures}
\label{appendix:supplementary}

This appendix contains the full numerical tables and per-dataset radar profiles that complement the main-body figures.

\begin{table}[H]
  \caption{Span-level precision and recall per PHI category for four LLMs on SHIELD in zero-shot extraction, with bootstrap 95\% CIs (2{,}000 document-level resamples). Support indicates the number of gold-standard spans per category.}
  \label{tab:llm-span}
  \centering
  \small
  \resizebox{\textwidth}{!}{%
  \begin{tabular}{llccccr}
    \toprule
    \textbf{Category} & \textbf{Metric} & \textbf{Gemini 2.5 Pro} & \textbf{Gemini 2.5 Flash} & \textbf{Llama 4 Maverick} & \textbf{GPT-OSS 120B} & \textbf{Support} \\
    \midrule
    AGE & Precision & 0.81 [0.71-0.89] & 0.85 [0.79-0.90] & 0.77 [0.67-0.85] & 0.83 [0.73-0.91] & 363 \\
    AGE & Recall & 0.91 [0.88-0.94] & 0.90 [0.87-0.93] & 0.79 [0.75-0.83] & 0.88 [0.85-0.92] &  \\
    DATE & Precision & 0.99 [0.98-0.99] & 0.99 [0.98-0.99] & 0.95 [0.94-0.97] & 0.98 [0.97-0.98] & 3547 \\
    DATE & Recall & 0.98 [0.98-0.99] & 0.98 [0.98-0.99] & 0.96 [0.95-0.97] & 0.97 [0.96-0.98] &  \\
    DOCTOR & Precision & 0.95 [0.94-0.96] & 0.95 [0.94-0.97] & 0.87 [0.85-0.89] & 0.96 [0.95-0.97] & 2598 \\
    DOCTOR & Recall & 0.97 [0.96-0.98] & 0.96 [0.96-0.97] & 0.84 [0.81-0.86] & 0.92 [0.90-0.93] &  \\
    HOSPITAL & Precision & 0.50 [0.47-0.54] & 0.67 [0.64-0.71] & 0.68 [0.64-0.71] & 0.62 [0.58-0.66] & 922 \\
    HOSPITAL & Recall & 0.91 [0.89-0.93] & 0.88 [0.85-0.90] & 0.82 [0.79-0.85] & 0.80 [0.77-0.84] &  \\
    ID & Precision & 0.89 [0.86-0.92] & 0.91 [0.87-0.94] & 0.91 [0.87-0.94] & 0.91 [0.87-0.94] & 728 \\
    ID & Recall & 0.95 [0.92-0.98] & 0.95 [0.92-0.97] & 0.80 [0.76-0.85] & 0.88 [0.84-0.91] &  \\
    LOCATION & Precision & 0.80 [0.75-0.85] & 0.86 [0.82-0.90] & 0.62 [0.55-0.68] & 0.85 [0.80-0.89] & 445 \\
    LOCATION & Recall & 0.66 [0.60-0.72] & 0.66 [0.60-0.71] & 0.60 [0.54-0.65] & 0.58 [0.52-0.64] &  \\
    PATIENT & Precision & 0.96 [0.95-0.97] & 0.95 [0.93-0.97] & 0.50 [0.46-0.55] & 0.90 [0.87-0.92] & 1031 \\
    PATIENT & Recall & 0.95 [0.94-0.97] & 0.95 [0.93-0.97] & 0.84 [0.80-0.87] & 0.89 [0.86-0.92] &  \\
    PHONE & Precision & 0.97 [0.95-0.98] & 0.98 [0.97-0.99] & 0.94 [0.91-0.96] & 0.98 [0.97-0.99] & 513 \\
    PHONE & Recall & 0.97 [0.95-0.99] & 0.95 [0.93-0.97] & 0.83 [0.78-0.88] & 0.96 [0.94-0.97] &  \\
    WEB & Precision & 0.97 [0.92-1.00] & 1.00 [1.00-1.00] & 0.94 [0.88-0.99] & 0.99 [0.95-1.00] & 82 \\
    WEB & Recall & 0.82 [0.67-0.95] & 0.82 [0.67-0.95] & 0.82 [0.67-0.95] & 0.82 [0.67-0.95] &  \\
    \bottomrule
  \end{tabular}}
\end{table}

\begin{table}[H]
  \caption{Span-level distillation comparison on SHIELD: Gemini 2.5 Flash (Teacher) vs.\ DeBERTa v3 (Student) with bootstrap 95\% CIs (2{,}000 document-level resamples). Support indicates the number of gold-standard spans per category.}
  \label{tab:distillation}
  \centering
  \small
  \resizebox{\textwidth}{!}{%
  \begin{tabular}{lccccr}
    \toprule
    & \multicolumn{2}{c}{\textbf{Gemini 2.5 Flash}} & \multicolumn{2}{c}{\textbf{DeBERTa v3}} & \\
    \cmidrule(lr){2-3} \cmidrule(lr){4-5}
    \textbf{Category} & \textbf{P [95\% CI]} & \textbf{R [95\% CI]} & \textbf{P [95\% CI]} & \textbf{R [95\% CI]} & \textbf{Support} \\
    \midrule
    AGE & 0.85 [0.79-0.90] & 0.90 [0.87-0.93] & 0.71 [0.65-0.77] & 0.78 [0.74-0.83] & 363 \\
    DATE & 0.99 [0.98-0.99] & 0.98 [0.98-0.99] & 0.97 [0.96-0.97] & 0.94 [0.93-0.95] & 3547 \\
    DOCTOR & 0.95 [0.94-0.97] & 0.96 [0.96-0.97] & 0.93 [0.92-0.94] & 0.90 [0.88-0.92] & 2598 \\
    HOSPITAL & 0.67 [0.64-0.71] & 0.88 [0.85-0.90] & 0.61 [0.57-0.65] & 0.79 [0.75-0.82] & 922 \\
    ID & 0.91 [0.87-0.94] & 0.95 [0.92-0.97] & 0.92 [0.89-0.94] & 0.90 [0.86-0.94] & 728 \\
    LOCATION & 0.86 [0.82-0.90] & 0.66 [0.60-0.71] & 0.78 [0.72-0.84] & 0.55 [0.49-0.61] & 445 \\
    PATIENT & 0.95 [0.93-0.97] & 0.95 [0.93-0.97] & 0.96 [0.94-0.97] & 0.88 [0.85-0.90] & 1031 \\
    PHONE & 0.98 [0.97-0.99] & 0.95 [0.93-0.97] & 0.89 [0.85-0.92] & 0.85 [0.81-0.89] & 513 \\
    WEB & 1.00 [1.00-1.00] & 0.82 [0.67-0.95] & 0.90 [0.81-0.98] & 0.67 [0.52-0.83] & 82 \\
    \bottomrule
  \end{tabular}}
\end{table}

\begin{table}[H]
  \caption{Span-level precision and recall of four transformer models on SHIELD with bootstrap 95\% CIs. Categories where AIMI models score 0.00 were not represented in their training data.}
  \label{tab:transformer-shield}
  \centering
  \small
  \resizebox{\textwidth}{!}{%
  \begin{tabular}{llcccc}
    \toprule
    \textbf{Category} & \textbf{Metric} & \textbf{AIMI v1} & \textbf{AIMI v2} & \textbf{BioModern} & \textbf{DeBERTa v3} \\
    \midrule
    AGE & Precision & 0.00 [0.00-0.00] & 0.00 [0.00-0.00] & 0.51 [0.47-0.56] & 0.71 [0.65-0.77] \\
    AGE & Recall & 0.00 [0.00-0.00] & 0.00 [0.00-0.00] & 0.76 [0.72-0.81] & 0.78 [0.74-0.83] \\
    DATE & Precision & 0.94 [0.93-0.95] & 0.95 [0.94-0.96] & 0.95 [0.94-0.96] & 0.97 [0.96-0.97] \\
    DATE & Recall & 0.95 [0.94-0.96] & 0.95 [0.94-0.96] & 0.95 [0.94-0.96] & 0.94 [0.93-0.95] \\
    DOCTOR & Precision & 0.41 [0.38-0.43] & 0.37 [0.34-0.39] & 0.91 [0.90-0.92] & 0.93 [0.92-0.94] \\
    DOCTOR & Recall & 0.46 [0.43-0.49] & 0.42 [0.40-0.45] & 0.87 [0.85-0.89] & 0.90 [0.88-0.92] \\
    HOSPITAL & Precision & 0.47 [0.44-0.50] & 0.39 [0.36-0.42] & 0.44 [0.41-0.48] & 0.61 [0.57-0.65] \\
    HOSPITAL & Recall & 0.90 [0.88-0.93] & 0.89 [0.87-0.91] & 0.72 [0.69-0.76] & 0.79 [0.75-0.82] \\
    ID & Precision & 0.63 [0.59-0.66] & 0.52 [0.49-0.56] & 0.70 [0.66-0.74] & 0.92 [0.89-0.94] \\
    ID & Recall & 0.95 [0.92-0.98] & 0.95 [0.92-0.97] & 0.86 [0.82-0.89] & 0.90 [0.86-0.94] \\
    LOCATION & Precision & 0.00 [0.00-0.00] & 0.00 [0.00-0.00] & 0.62 [0.56-0.69] & 0.78 [0.72-0.84] \\
    LOCATION & Recall & 0.00 [0.00-0.00] & 0.00 [0.00-0.00] & 0.56 [0.50-0.61] & 0.55 [0.49-0.61] \\
    PATIENT & Precision & 0.88 [0.86-0.91] & 0.89 [0.86-0.91] & 0.85 [0.81-0.88] & 0.96 [0.94-0.97] \\
    PATIENT & Recall & 0.78 [0.75-0.81] & 0.71 [0.67-0.75] & 0.81 [0.78-0.84] & 0.88 [0.85-0.90] \\
    PHONE & Precision & 0.69 [0.64-0.73] & 0.76 [0.72-0.80] & 0.75 [0.69-0.81] & 0.89 [0.85-0.92] \\
    PHONE & Recall & 0.92 [0.89-0.94] & 0.81 [0.77-0.86] & 0.83 [0.79-0.87] & 0.85 [0.81-0.89] \\
    WEB & Precision & 0.00 [0.00-0.00] & 0.00 [0.00-0.00] & 0.43 [0.31-0.54] & 0.90 [0.81-0.98] \\
    WEB & Recall & 0.00 [0.00-0.00] & 0.00 [0.00-0.00] & 0.67 [0.53-0.82] & 0.67 [0.52-0.83] \\
    \bottomrule
  \end{tabular}}
\end{table}

\begin{table}[H]
  \caption{Span-level precision and recall of four transformer models on i2b2 2014 (cross-dataset) with bootstrap 95\% CIs. AGE is near-zero across all models due to label-definition mismatch. WEB has very low support (only 5 gold spans), so its estimates are reported but should be interpreted with caution.}
  \label{tab:cross-i2b2}
  \centering
  \small
  \resizebox{\textwidth}{!}{%
  \begin{tabular}{llcccc}
    \toprule
    \textbf{Category} & \textbf{Metric} & \textbf{AIMI v1} & \textbf{AIMI v2} & \textbf{BioModern} & \textbf{DeBERTa v3} \\
    \midrule
    AGE & Precision & 0.00 [0.00-0.00] & 0.14 [0.03-0.28] & 0.03 [0.02-0.03] & 0.01 [0.00-0.01] \\
    AGE & Recall & 0.00 [0.00-0.00] & 0.00 [0.00-0.00] & 0.05 [0.03-0.06] & 0.01 [0.00-0.01] \\
    DATE & Precision & 0.98 [0.98-0.99] & 0.98 [0.97-0.98] & 0.93 [0.92-0.93] & 0.96 [0.95-0.96] \\
    DATE & Recall & 0.94 [0.93-0.95] & 0.94 [0.93-0.94] & 0.86 [0.85-0.87] & 0.85 [0.84-0.86] \\
    DOCTOR & Precision & 0.25 [0.23-0.27] & 0.23 [0.21-0.25] & 0.86 [0.85-0.88] & 0.92 [0.91-0.93] \\
    DOCTOR & Recall & 0.25 [0.23-0.27] & 0.23 [0.21-0.25] & 0.85 [0.83-0.86] & 0.85 [0.84-0.87] \\
    HOSPITAL & Precision & 0.59 [0.57-0.61] & 0.52 [0.49-0.54] & 0.39 [0.37-0.41] & 0.48 [0.45-0.51] \\
    HOSPITAL & Recall & 0.87 [0.85-0.88] & 0.84 [0.82-0.86] & 0.44 [0.42-0.47] & 0.39 [0.37-0.42] \\
    ID & Precision & 0.83 [0.81-0.85] & 0.73 [0.71-0.75] & 0.52 [0.49-0.54] & 0.85 [0.82-0.87] \\
    ID & Recall & 0.81 [0.75-0.87] & 0.80 [0.74-0.86] & 0.66 [0.60-0.71] & 0.64 [0.58-0.69] \\
    LOCATION & Precision & 0.00 [0.00-0.00] & 0.00 [0.00-0.00] & 0.36 [0.33-0.39] & 0.49 [0.45-0.52] \\
    LOCATION & Recall & 0.00 [0.00-0.00] & 0.00 [0.00-0.00] & 0.22 [0.20-0.24] & 0.26 [0.23-0.28] \\
    PATIENT & Precision & 0.55 [0.51-0.58] & 0.53 [0.50-0.56] & 0.58 [0.54-0.61] & 0.90 [0.88-0.93] \\
    PATIENT & Recall & 0.56 [0.53-0.60] & 0.54 [0.51-0.57] & 0.63 [0.60-0.66] & 0.79 [0.77-0.82] \\
    PHONE & Precision & 0.76 [0.72-0.80] & 0.77 [0.73-0.81] & 0.21 [0.19-0.24] & 0.73 [0.68-0.78] \\
    PHONE & Recall & 0.94 [0.92-0.96] & 0.88 [0.85-0.91] & 0.51 [0.47-0.56] & 0.56 [0.52-0.60] \\
    WEB & Precision & 0.00 [0.00-0.00] & 0.00 [0.00-0.00] & 0.01 [0.00-0.02] & 0.22 [0.05-0.47] \\
    WEB & Recall & 0.00 [0.00-0.00] & 0.00 [0.00-0.00] & 1.00 [1.00-1.00] & 0.80 [0.28-1.00] \\
    \bottomrule
  \end{tabular}}
\end{table}

\begin{table}[H]
  \caption{Span-level precision and recall of four transformer models on AIMI (cross-dataset) with bootstrap 95\% CIs. WEB is excluded (0 gold spans in AIMI). LOCATION and PATIENT have very low support.}
  \label{tab:cross-aimi}
  \centering
  \small
  \resizebox{\textwidth}{!}{%
  \begin{tabular}{llcccc}
    \toprule
    \textbf{Category} & \textbf{Metric} & \textbf{AIMI v1} & \textbf{AIMI v2} & \textbf{BioModern} & \textbf{DeBERTa v3} \\
    \midrule
    AGE & Precision & 0.00 [0.00-0.00] & 0.99 [0.98-0.99] & 0.00 [0.00-0.00] & 0.00 [0.00-0.00] \\
    AGE & Recall & 0.00 [0.00-0.00] & 0.98 [0.97-0.99] & 0.00 [0.00-0.00] & 0.00 [0.00-0.00] \\
    DATE & Precision & 0.98 [0.98-0.98] & 0.98 [0.98-0.98] & 0.96 [0.96-0.96] & 0.98 [0.98-0.98] \\
    DATE & Recall & 0.96 [0.96-0.96] & 0.96 [0.96-0.96] & 0.93 [0.93-0.94] & 0.94 [0.94-0.95] \\
    DOCTOR & Precision & 0.99 [0.99-0.99] & 1.00 [1.00-1.00] & 0.98 [0.97-0.98] & 0.99 [0.99-0.99] \\
    DOCTOR & Recall & 0.99 [0.99-1.00] & 1.00 [1.00-1.00] & 0.98 [0.98-0.98] & 0.99 [0.99-0.99] \\
    HOSPITAL & Precision & 0.81 [0.77-0.84] & 0.94 [0.92-0.95] & 0.00 [0.00-0.01] & 0.06 [0.04-0.08] \\
    HOSPITAL & Recall & 0.36 [0.33-0.39] & 0.99 [0.99-1.00] & 0.09 [0.07-0.10] & 0.02 [0.01-0.03] \\
    ID & Precision & 1.00 [1.00-1.00] & 1.00 [1.00-1.00] & 0.95 [0.95-0.96] & 0.97 [0.97-0.97] \\
    ID & Recall & 1.00 [0.99-1.00] & 1.00 [1.00-1.00] & 0.84 [0.84-0.85] & 0.96 [0.95-0.96] \\
    LOCATION & Precision & 0.00 [0.00-0.00] & 0.00 [0.00-0.00] & 0.00 [0.00-0.00] & 0.05 [0.00-0.14] \\
    LOCATION & Recall & 0.00 [0.00-0.00] & 0.00 [0.00-0.00] & 0.00 [0.00-0.00] & 0.02 [0.00-0.05] \\
    PATIENT & Precision & 0.16 [0.10-0.23] & 0.87 [0.71-1.00] & 0.01 [0.00-0.02] & 0.20 [0.09-0.32] \\
    PATIENT & Recall & 0.78 [0.61-0.93] & 0.74 [0.56-0.90] & 0.37 [0.18-0.57] & 0.44 [0.25-0.65] \\
    PHONE & Precision & 0.98 [0.98-0.99] & 1.00 [1.00-1.00] & 0.37 [0.36-0.38] & 0.94 [0.93-0.95] \\
    PHONE & Recall & 0.99 [0.99-1.00] & 1.00 [0.99-1.00] & 0.99 [0.98-0.99] & 0.99 [0.99-0.99] \\
    WEB & Precision &  &  &  &  \\
    WEB & Recall &  &  &  &  \\
    \bottomrule
  \end{tabular}}
\end{table}

\begin{figure}[H]
  \centering
  \includegraphics[width=\textwidth,alt={Radar plots of span-level performance for four transformer models evaluated cross-dataset on i2b2 2014; DeBERTa v3 achieves the most balanced profile.}]{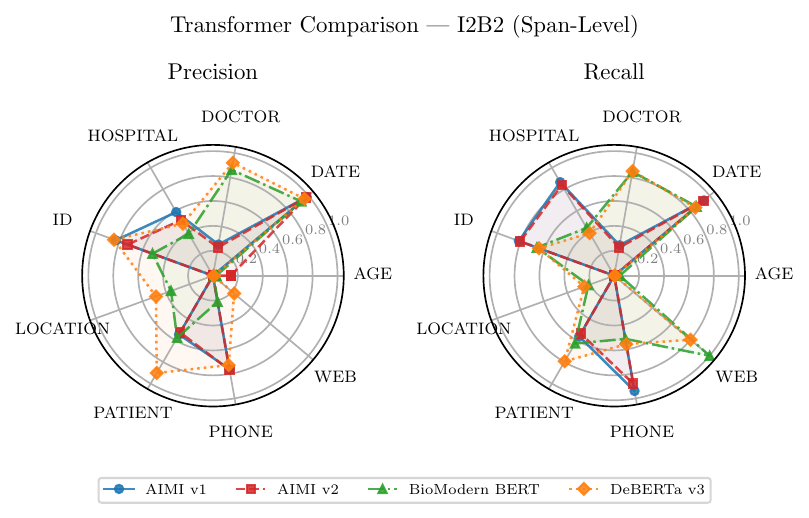}
  \caption{Span-level radar comparison of four transformer models on i2b2 2014 (cross-dataset). DeBERTa v3 achieves the most balanced profile, while AIMI v1/v2 show strong DATE recall but lack coverage on LOCATION.}
  \label{fig:radar-transformer-i2b2}
\end{figure}

\begin{figure}[H]
  \centering
  \includegraphics[width=\textwidth,alt={Radar plots of span-level performance for four transformer models evaluated cross-dataset on AIMI; AIMI v2 dominates on its home data while DeBERTa v3 and BioClinical ModernBERT show competitive recall on universal categories.}]{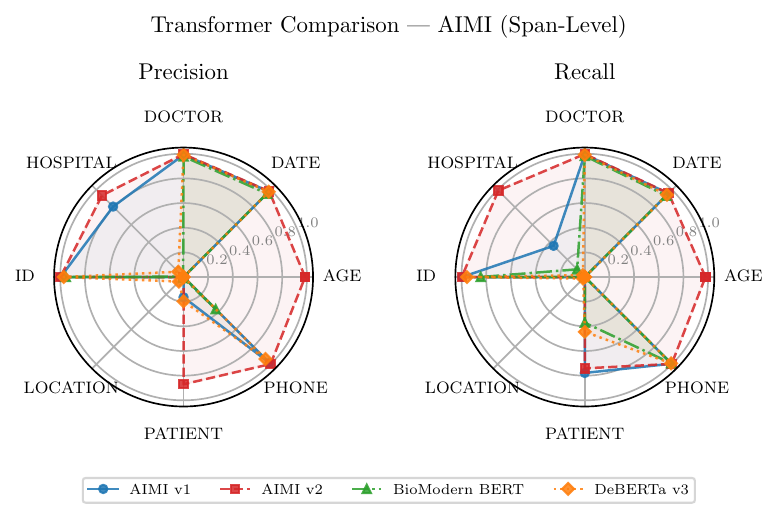}
  \caption{Span-level radar comparison of four transformer models on AIMI (cross-dataset). AIMI v2 dominates on its home data (near-perfect on DOCTOR, ID, HOSPITAL), while DeBERTa v3 and BioModern BERT show competitive recall on universal categories but limited performance on institution-specific HOSPITAL.}
  \label{fig:radar-transformer-aimi}
\end{figure}

\section{Unified Label Taxonomy Mappings}
\label{appendix:label-mappings}

To enable cross-dataset evaluation, all gold-standard annotations and model outputs were mapped into a unified canonical label taxonomy consisting of 10 categories: \texttt{DATE}, \texttt{ID}, \texttt{DOCTOR}, \texttt{PHONE}, \texttt{AGE}, \texttt{HOSPITAL}, \texttt{LOCATION}, \texttt{PATIENT}, \texttt{WEB}, and \texttt{OTHER}. The \texttt{OTHER} category (to which i2b2 \texttt{PROFESSION} and \texttt{ORGANIZATION} are mapped) is a catch-all that is \emph{excluded from all scoring}; it therefore appears in no results table, and every reported precision/recall figure is computed over the \shieldNumCats{} scored categories. Tables~\ref{tab:mapping-i2b2} and~\ref{tab:mapping-aimi} define the complete mappings used in all experiments reported in this paper.

\begin{table}[H]
  \caption{i2b2 2014 to unified (SHIELD) label mapping. Multiple fine-grained i2b2 categories are consolidated into broader canonical categories.}
  \label{tab:mapping-i2b2}
  \centering
  \small
  \begin{tabular}{ll}
    \toprule
    \textbf{i2b2 Label} & \textbf{Unified Label} \\
    \midrule
    DATE            & DATE \\
    PATIENT         & PATIENT \\
    DOCTOR          & DOCTOR \\
    MEDICALRECORD   & ID \\
    IDNUM           & ID \\
    USERNAME        & ID \\
    DEVICE          & ID \\
    AGE             & AGE \\
    HOSPITAL        & HOSPITAL \\
    PHONE           & PHONE \\
    FAX             & PHONE \\
    STREET          & LOCATION \\
    CITY            & LOCATION \\
    STATE           & LOCATION \\
    ZIP             & LOCATION \\
    COUNTRY         & LOCATION \\
    LOCATION-OTHER  & LOCATION \\
    EMAIL           & WEB \\
    PROFESSION      & OTHER \\
    ORGANIZATION    & OTHER \\
    \bottomrule
  \end{tabular}
\end{table}

\begin{table}[H]
  \caption{AIMI to unified (SHIELD) label mapping. Note that AIMI's \texttt{HOSPITAL} maps to \texttt{LOCATION} and \texttt{VENDOR} maps to \texttt{HOSPITAL} in the unified taxonomy, reflecting semantic differences in annotation guidelines between datasets.}
  \label{tab:mapping-aimi}
  \centering
  \small
  \begin{tabular}{ll}
    \toprule
    \textbf{AIMI Label} & \textbf{Unified Label} \\
    \midrule
    DATES    & DATE \\
    PATIENT  & PATIENT \\
    HCW      & DOCTOR \\
    UNIQUE   & ID \\
    HOSPITAL & LOCATION \\
    VENDOR   & HOSPITAL \\
    PHONE    & PHONE \\
    AGE      & AGE \\
    \bottomrule
  \end{tabular}
\end{table}

\section{Token-Level Evaluation Results}
\label{appendix:token-level}

Section~\ref{sec:distillation-results} and the main results tables report span-level metrics (see Section~\ref{sec:statistical-analysis} for the matching criterion). This appendix presents the complementary \emph{token-level} evaluation, where each token is independently classified and scored. Token-level metrics can diverge meaningfully from span-level metrics. For example, a model may correctly tag most tokens within a span but miss the boundary, yielding high token-level recall with lower span-level recall.

All tables below are generated automatically from the evaluation pipeline and read directly from its CSV outputs, which keeps the reported numbers consistent with the pipeline results.

\subsection{LLM Token-Level Performance on SHIELD}

\begin{table}[H]
  \caption{Token-level precision and recall for four LLMs on SHIELD, with bootstrap 95\% CIs (2{,}000 document-level resamples). Support indicates the number of gold-standard tokens per category.}
  \label{tab:token-llm-shield}
  \centering
  \small
  \resizebox{\textwidth}{!}{%
  \begin{tabular}{llccccr}
    \toprule
    \textbf{Category} & \textbf{Metric} & \textbf{Gemini 2.5 Pro} & \textbf{Gemini 2.5 Flash} & \textbf{Llama 4 Maverick} & \textbf{GPT-OSS 120B} & \textbf{Support} \\
    \midrule
    AGE & Precision & 0.87 [0.81-0.92] & 0.90 [0.85-0.93] & 0.86 [0.80-0.90] & 0.90 [0.85-0.94] & 1033 \\
    AGE & Recall & 0.96 [0.94-0.98] & 0.96 [0.93-0.98] & 0.85 [0.80-0.88] & 0.93 [0.90-0.96] &  \\
    DATE & Precision & 0.99 [0.99-1.00] & 0.99 [0.99-1.00] & 0.98 [0.97-0.98] & 0.99 [0.99-0.99] & 17401 \\
    DATE & Recall & 0.99 [0.99-0.99] & 0.99 [0.99-1.00] & 0.99 [0.98-0.99] & 0.99 [0.98-0.99] &  \\
    DOCTOR & Precision & 0.83 [0.82-0.84] & 0.92 [0.91-0.93] & 0.87 [0.85-0.88] & 0.90 [0.89-0.91] & 8645 \\
    DOCTOR & Recall & 0.99 [0.98-0.99] & 0.98 [0.97-0.98] & 0.92 [0.90-0.93] & 0.94 [0.93-0.96] &  \\
    HOSPITAL & Precision & 0.53 [0.50-0.56] & 0.71 [0.67-0.74] & 0.73 [0.70-0.77] & 0.64 [0.60-0.68] & 3303 \\
    HOSPITAL & Recall & 0.96 [0.93-0.98] & 0.91 [0.89-0.94] & 0.86 [0.83-0.89] & 0.87 [0.84-0.90] &  \\
    ID & Precision & 0.93 [0.90-0.96] & 0.94 [0.92-0.96] & 0.94 [0.91-0.96] & 0.93 [0.91-0.96] & 4115 \\
    ID & Recall & 0.95 [0.91-0.97] & 0.94 [0.90-0.97] & 0.78 [0.73-0.82] & 0.83 [0.79-0.87] &  \\
    LOCATION & Precision & 0.94 [0.92-0.96] & 0.95 [0.93-0.97] & 0.86 [0.83-0.89] & 0.94 [0.92-0.96] & 4100 \\
    LOCATION & Recall & 0.90 [0.88-0.93] & 0.90 [0.88-0.92] & 0.82 [0.78-0.85] & 0.85 [0.81-0.88] &  \\
    PATIENT & Precision & 0.96 [0.94-0.97] & 0.95 [0.93-0.97] & 0.67 [0.63-0.71] & 0.91 [0.88-0.93] & 2995 \\
    PATIENT & Recall & 0.96 [0.94-0.97] & 0.95 [0.93-0.97] & 0.90 [0.87-0.92] & 0.93 [0.90-0.95] &  \\
    PHONE & Precision & 0.99 [0.98-0.99] & 0.99 [0.98-1.00] & 0.97 [0.95-0.98] & 0.99 [0.99-1.00] & 3605 \\
    PHONE & Recall & 0.98 [0.97-0.99] & 0.97 [0.95-0.98] & 0.80 [0.74-0.86] & 0.97 [0.95-0.98] &  \\
    WEB & Precision & 0.99 [0.97-1.00] & 1.00 [1.00-1.00] & 0.98 [0.95-1.00] & 0.99 [0.95-1.00] & 602 \\
    WEB & Recall & 0.88 [0.75-0.97] & 0.88 [0.75-0.97] & 0.88 [0.75-0.97] & 0.88 [0.75-0.97] &  \\
    \bottomrule
  \end{tabular}}
\end{table}

\subsection{Token-Level Distillation Comparison on SHIELD}

\begin{table}[H]
  \caption{Token-level distillation comparison on SHIELD: Gemini 2.5 Flash (Teacher) vs.\ DeBERTa v3 (Student) with bootstrap 95\% CIs (2{,}000 document-level resamples). Support indicates the number of gold-standard tokens per category.}
  \label{tab:token-distillation-shield}
  \centering
  \small
  \resizebox{\textwidth}{!}{%
  \begin{tabular}{lccccr}
    \toprule
    & \multicolumn{2}{c}{\textbf{Gemini 2.5 Flash}} & \multicolumn{2}{c}{\textbf{DeBERTa v3}} & \\
    \cmidrule(lr){2-3} \cmidrule(lr){4-5}
    \textbf{Category} & \textbf{P [95\% CI]} & \textbf{R [95\% CI]} & \textbf{P [95\% CI]} & \textbf{R [95\% CI]} & \textbf{Support} \\
    \midrule
    AGE & 0.90 [0.85-0.93] & 0.96 [0.93-0.98] & 0.83 [0.78-0.87] & 0.98 [0.96-0.99] & 1033 \\
    DATE & 0.99 [0.99-1.00] & 0.99 [0.99-1.00] & 0.98 [0.98-0.99] & 0.99 [0.99-0.99] & 17401 \\
    DOCTOR & 0.92 [0.91-0.93] & 0.98 [0.97-0.98] & 0.87 [0.86-0.88] & 0.97 [0.96-0.98] & 8645 \\
    HOSPITAL & 0.71 [0.67-0.74] & 0.91 [0.89-0.94] & 0.66 [0.62-0.70] & 0.88 [0.85-0.91] & 3303 \\
    ID & 0.94 [0.92-0.96] & 0.94 [0.90-0.97] & 0.94 [0.92-0.96] & 0.93 [0.90-0.96] & 4115 \\
    LOCATION & 0.95 [0.93-0.97] & 0.90 [0.88-0.92] & 0.91 [0.88-0.93] & 0.87 [0.84-0.90] & 4100 \\
    PATIENT & 0.95 [0.93-0.97] & 0.95 [0.93-0.97] & 0.97 [0.95-0.98] & 0.90 [0.88-0.92] & 2995 \\
    PHONE & 0.99 [0.98-1.00] & 0.97 [0.95-0.98] & 0.93 [0.90-0.95] & 0.95 [0.92-0.97] & 3605 \\
    WEB & 1.00 [1.00-1.00] & 0.88 [0.75-0.97] & 0.95 [0.87-1.00] & 0.86 [0.74-0.96] & 602 \\
    \bottomrule
  \end{tabular}}
\end{table}

\subsection{Transformer Token-Level Performance on SHIELD}

\begin{table}[H]
  \caption{Token-level precision and recall of four transformer models on SHIELD, with bootstrap 95\% CIs (2{,}000 document-level resamples). Categories where AIMI models score 0.00 were not represented in their training data.}
  \label{tab:token-transformer-shield}
  \centering
  \small
  \resizebox{\textwidth}{!}{%
  \begin{tabular}{llccccr}
    \toprule
    \textbf{Category} & \textbf{Metric} & \textbf{AIMI v1} & \textbf{AIMI v2} & \textbf{BioModern} & \textbf{DeBERTa v3} & \textbf{Support} \\
    \midrule
    AGE & Precision & 0.00 [0.00-0.00] & 0.00 [0.00-0.00] & 0.69 [0.65-0.73] & 0.83 [0.78-0.87] & 1033 \\
    AGE & Recall & 0.00 [0.00-0.00] & 0.00 [0.00-0.00] & 0.97 [0.95-0.98] & 0.98 [0.96-0.99] &  \\
    DATE & Precision & 0.98 [0.97-0.98] & 0.98 [0.97-0.98] & 0.98 [0.97-0.98] & 0.98 [0.98-0.99] & 17401 \\
    DATE & Recall & 0.99 [0.98-0.99] & 0.99 [0.98-0.99] & 0.99 [0.99-0.99] & 0.99 [0.99-0.99] &  \\
    DOCTOR & Precision & 0.59 [0.58-0.60] & 0.56 [0.55-0.58] & 0.88 [0.87-0.90] & 0.87 [0.86-0.88] & 8645 \\
    DOCTOR & Recall & 0.98 [0.97-0.99] & 0.98 [0.97-0.99] & 0.94 [0.93-0.95] & 0.97 [0.96-0.98] &  \\
    HOSPITAL & Precision & 0.46 [0.44-0.49] & 0.43 [0.40-0.45] & 0.58 [0.54-0.61] & 0.66 [0.62-0.70] & 3303 \\
    HOSPITAL & Recall & 0.93 [0.92-0.95] & 0.95 [0.93-0.96] & 0.84 [0.81-0.87] & 0.88 [0.85-0.91] &  \\
    ID & Precision & 0.75 [0.72-0.78] & 0.64 [0.60-0.67] & 0.87 [0.84-0.89] & 0.94 [0.92-0.96] & 4115 \\
    ID & Recall & 0.96 [0.95-0.98] & 0.99 [0.98-1.00] & 0.82 [0.78-0.86] & 0.93 [0.90-0.96] &  \\
    LOCATION & Precision & 0.00 [0.00-0.00] & 0.00 [0.00-0.00] & 0.89 [0.86-0.92] & 0.91 [0.88-0.93] & 4100 \\
    LOCATION & Recall & 0.00 [0.00-0.00] & 0.00 [0.00-0.00] & 0.85 [0.82-0.88] & 0.87 [0.84-0.90] &  \\
    PATIENT & Precision & 0.91 [0.89-0.93] & 0.92 [0.90-0.94] & 0.89 [0.87-0.92] & 0.97 [0.95-0.98] & 2995 \\
    PATIENT & Recall & 0.84 [0.81-0.87] & 0.76 [0.73-0.80] & 0.86 [0.83-0.89] & 0.90 [0.88-0.92] &  \\
    PHONE & Precision & 0.86 [0.84-0.89] & 0.90 [0.87-0.92] & 0.88 [0.85-0.91] & 0.93 [0.90-0.95] & 3605 \\
    PHONE & Recall & 0.95 [0.93-0.97] & 0.83 [0.78-0.87] & 0.94 [0.91-0.96] & 0.95 [0.92-0.97] &  \\
    WEB & Precision & 0.00 [0.00-0.00] & 0.00 [0.00-0.00] & 0.83 [0.74-0.89] & 0.95 [0.87-1.00] & 602 \\
    WEB & Recall & 0.00 [0.00-0.00] & 0.00 [0.00-0.00] & 0.86 [0.74-0.95] & 0.86 [0.74-0.96] &  \\
    \bottomrule
  \end{tabular}}
\end{table}

\subsection{Transformer Token-Level Performance on i2b2 (Cross-Dataset)}

\begin{table}[H]
  \caption{Token-level transformer performance on i2b2 2014 (cross-dataset), with bootstrap 95\% CIs (2{,}000 document-level resamples).}
  \label{tab:token-transformer-i2b2}
  \centering
  \small
  \resizebox{\textwidth}{!}{%
  \begin{tabular}{llccccr}
    \toprule
    \textbf{Category} & \textbf{Metric} & \textbf{AIMI v1} & \textbf{AIMI v2} & \textbf{BioModern} & \textbf{DeBERTa v3} & \textbf{Support} \\
    \midrule
    AGE & Precision & 0.00 [0.00-0.00] & 0.14 [0.03-0.28] & 0.25 [0.24-0.26] & 0.34 [0.33-0.35] & 2149 \\
    AGE & Recall & 0.00 [0.00-0.00] & 0.00 [0.00-0.00] & 0.90 [0.88-0.92] & 0.86 [0.84-0.89] &  \\
    DATE & Precision & 1.00 [0.99-1.00] & 0.99 [0.99-0.99] & 0.98 [0.98-0.98] & 0.98 [0.98-0.99] & 55575 \\
    DATE & Recall & 0.97 [0.96-0.97] & 0.97 [0.96-0.97] & 0.95 [0.94-0.95] & 0.95 [0.95-0.96] &  \\
    DOCTOR & Precision & 0.55 [0.55-0.56] & 0.53 [0.52-0.54] & 0.81 [0.79-0.83] & 0.82 [0.80-0.83] & 13618 \\
    DOCTOR & Recall & 0.98 [0.97-0.98] & 0.98 [0.97-0.98] & 0.93 [0.93-0.94] & 0.94 [0.93-0.95] &  \\
    HOSPITAL & Precision & 0.58 [0.56-0.61] & 0.55 [0.53-0.57] & 0.63 [0.61-0.65] & 0.71 [0.69-0.73] & 6460 \\
    HOSPITAL & Recall & 0.96 [0.95-0.97] & 0.96 [0.95-0.97] & 0.74 [0.72-0.77] & 0.73 [0.70-0.76] &  \\
    ID & Precision & 0.90 [0.89-0.91] & 0.83 [0.81-0.84] & 0.75 [0.73-0.77] & 0.91 [0.89-0.92] & 9273 \\
    ID & Recall & 0.89 [0.84-0.93] & 0.89 [0.84-0.93] & 0.72 [0.68-0.76] & 0.72 [0.67-0.76] &  \\
    LOCATION & Precision & 0.00 [0.00-0.00] & 0.00 [0.00-0.00] & 0.74 [0.72-0.76] & 0.70 [0.68-0.72] & 4668 \\
    LOCATION & Recall & 0.00 [0.00-0.00] & 0.00 [0.00-0.00] & 0.80 [0.77-0.83] & 0.83 [0.80-0.86] &  \\
    PATIENT & Precision & 0.71 [0.70-0.73] & 0.70 [0.68-0.71] & 0.76 [0.74-0.78] & 0.94 [0.92-0.95] & 6253 \\
    PATIENT & Recall & 0.96 [0.94-0.97] & 0.93 [0.91-0.95] & 0.80 [0.78-0.83] & 0.82 [0.80-0.84] &  \\
    PHONE & Precision & 0.85 [0.82-0.88] & 0.85 [0.82-0.87] & 0.45 [0.40-0.49] & 0.80 [0.76-0.83] & 2613 \\
    PHONE & Recall & 0.97 [0.96-0.99] & 0.91 [0.88-0.94] & 0.77 [0.73-0.80] & 0.81 [0.78-0.85] &  \\
    WEB & Precision & 0.00 [0.00-0.00] & 0.00 [0.00-0.00] & 0.06 [0.01-0.11] & 0.34 [0.09-0.60] & 46 \\
    WEB & Recall & 0.00 [0.00-0.00] & 0.00 [0.00-0.00] & 1.00 [1.00-1.00] & 1.00 [1.00-1.00] &  \\
    \bottomrule
  \end{tabular}}
\end{table}

\subsection{Transformer Token-Level Performance on AIMI (Cross-Dataset)}

\begin{table}[H]
  \caption{Token-level transformer performance on AIMI (cross-dataset), with bootstrap 95\% CIs (2{,}000 document-level resamples). WEB is excluded (0 gold tokens in AIMI).}
  \label{tab:token-transformer-aimi}
  \centering
  \small
  \resizebox{\textwidth}{!}{%
  \begin{tabular}{llccccr}
    \toprule
    \textbf{Category} & \textbf{Metric} & \textbf{AIMI v1} & \textbf{AIMI v2} & \textbf{BioModern} & \textbf{DeBERTa v3} & \textbf{Support} \\
    \midrule
    AGE & Precision & 0.00 [0.00-0.00] & 0.99 [0.98-0.99] & 0.01 [0.01-0.01] & 0.01 [0.01-0.01] & 1033 \\
    AGE & Recall & 0.00 [0.00-0.00] & 0.98 [0.96-0.99] & 0.96 [0.95-0.98] & 0.97 [0.96-0.98] &  \\
    DATE & Precision & 0.99 [0.99-0.99] & 0.99 [0.99-1.00] & 0.99 [0.99-0.99] & 0.99 [0.99-0.99] & 507934 \\
    DATE & Recall & 1.00 [1.00-1.00] & 1.00 [1.00-1.00] & 0.99 [0.99-0.99] & 0.99 [0.99-0.99] &  \\
    DOCTOR & Precision & 0.99 [0.99-1.00] & 1.00 [1.00-1.00] & 0.97 [0.96-0.97] & 0.98 [0.98-0.98] & 52137 \\
    DOCTOR & Recall & 0.99 [0.99-0.99] & 1.00 [1.00-1.00] & 0.79 [0.79-0.80] & 0.76 [0.76-0.77] &  \\
    HOSPITAL & Precision & 0.82 [0.79-0.86] & 0.94 [0.92-0.95] & 0.09 [0.08-0.09] & 0.43 [0.42-0.45] & 6925 \\
    HOSPITAL & Recall & 0.31 [0.28-0.34] & 0.99 [0.99-1.00] & 0.61 [0.58-0.64] & 0.33 [0.29-0.36] &  \\
    ID & Precision & 1.00 [1.00-1.00] & 1.00 [1.00-1.00] & 0.99 [0.99-0.99] & 0.99 [0.99-0.99] & 285974 \\
    ID & Recall & 1.00 [1.00-1.00] & 1.00 [1.00-1.00] & 0.79 [0.78-0.79] & 0.97 [0.97-0.97] &  \\
    LOCATION & Precision & 0.00 [0.00-0.00] & 0.00 [0.00-0.00] & 0.00 [0.00-0.00] & 0.01 [0.00-0.03] & 398 \\
    LOCATION & Recall & 0.00 [0.00-0.00] & 0.00 [0.00-0.00] & 0.00 [0.00-0.01] & 0.01 [0.00-0.01] &  \\
    PATIENT & Precision & 0.16 [0.09-0.23] & 0.87 [0.70-1.00] & 0.01 [0.00-0.01] & 0.13 [0.06-0.21] & 76 \\
    PATIENT & Recall & 0.82 [0.66-0.94] & 0.76 [0.59-0.92] & 0.43 [0.24-0.62] & 0.42 [0.23-0.62] &  \\
    PHONE & Precision & 0.99 [0.99-1.00] & 1.00 [1.00-1.00] & 0.37 [0.35-0.38] & 0.94 [0.93-0.95] & 17464 \\
    PHONE & Recall & 1.00 [0.99-1.00] & 1.00 [1.00-1.00] & 0.99 [0.99-1.00] & 0.99 [0.99-1.00] &  \\
    WEB & Precision &  &  &  &  &  \\
    WEB & Recall &  &  &  &  &  \\
    \bottomrule
  \end{tabular}}
\end{table}

\section{LLM Extraction Prompt}
\label{appendix:llm-prompt}

The following prompt was used for all four LLMs (Gemini 2.5 Pro, Gemini 2.5 Flash, Llama 4 Maverick, GPT-OSS 120B) in the zero-shot extraction evaluation (Figure~\ref{fig:llm-benchmark}) and for the teacher labeling stage of the distillation pipeline (Section~\ref{sec:model-selection}). The placeholder \texttt{\{clinical\_text\}} is replaced with the raw clinical note at inference time. The prompt was iteratively refined on a small calibration subset of the SHIELD gold standard (not used for student model training).

\begin{tcolorbox}[breakable, colback=white, colframe=black, boxrule=0.5pt, left=6pt, right=6pt, top=6pt, bottom=6pt]
\small
\textit{Identify and tag ALL patient-identifiable information and organizational information from clinical notes. Return results as JSON.}

\medskip
\textbf{TOP PRIORITIES (NEVER MISS THESE)}

\begin{enumerate}[leftmargin=*]
    \item \textbf{DATES}: Scan exhaustively for ALL date formats (MM/DD/YYYY, M/D/YY, M/D, Month DD YYYY, etc.). Include short dates like ``4/12'' without years. Tag every occurrence separately.
    \begin{itemize}[leftmargin=*]
        \item EXCLUDE times (10:00AM), temporal words (yesterday), day names (Monday). Example: ``2023-03-05T03:43:00-08:00'' $\rightarrow$ ``2023-03-05'' (exclude T03:43:00-08:00).
    \end{itemize}
    \item \textbf{COMPLETE ADDRESSES WITH ZIP}: ALWAYS tag addresses in format ``Street Address \;\; City State ZIP'' (note: 2+ spaces between street and city). Example: ``1234 MAIN STREET \;\; BOSTON MA 02101'' is ONE complete entity.
    \item \textbf{ALL DISTINCT MENTIONS}: Tag each occurrence separately. Do NOT skip any mention because another version exists elsewhere.
    \item \textbf{EXACT TEXT MATCHING}: Copy text exactly---preserve all case, whitespace, punctuation. Different whitespace = different entity.
\end{enumerate}

\medskip
\textbf{ENTITY DEFINITIONS}

\begin{itemize}[leftmargin=*]
    \item \textbf{PATIENT}: Patient names, relatives, friends, guardians, and any person name that is NOT a healthcare provider. EXCLUDE generic terms (``sister'', ``mother'') and titles (Ms./Mr./Mrs., PhD).
    \item \textbf{LOCATION}: ALL address mentions (complete or incomplete). Preserve internal whitespace exactly. EXCLUDE generic facility names (PACU, ICU) without room numbers.
    \item \textbf{DATE}: Any date format. Past, present, future.
    \item \textbf{ID}: MRNs (include hyphens: ``1234567-8''), SSNs, account numbers. EXCLUDE generic codes (R07.89), labels without values (``MRN'').
    \item \textbf{AGE}: Complete expressions with units (``72 Y old'', ``14 year old''). When units in separate fields (``Age: 14 \;\; Units: Years''), tag only number ``14''.
    \item \textbf{PHONE}: All formats including pagers. For pagers, tag ONLY number (``12019''), NOT label (``Pgr'').
    \item \textbf{WEB}: Emails, URLs, social media handles, IPs.
    \item \textbf{DOCTOR}: ALL healthcare worker names (providers, nurses, staff, interpreters, coordinators, financial staff). Formal: ``Dr.\ Smith'' $\rightarrow$ ``Smith''; ``Wilson, Anne Marie, MD'' $\rightarrow$ ``Wilson, Anne Marie,''. Informal first names when referring to healthcare workers. EXCLUDE trailing titles; INCLUDE mid-name titles.
    \item \textbf{HOSPITAL}: Organizations, pharmacies, clinics. Room numbers as number only (``3142'', NOT ``room 3142''). EXCLUDE generic teams (TT MICU GREEN), generic services (language line).
    \item \textbf{OTHER}: Unique identifying characteristics.
\end{itemize}

\medskip
\textbf{KEY RULES}

\begin{itemize}[leftmargin=*]
    \item \textbf{Organization + Location Combinations}: Split into HOSPITAL + LOCATION components. Shorthand (org + city/state only): split into separate entities.
    \item \textbf{Multiple Address Formats}: When the same physical location appears in different formats, tag EACH format separately.
    \item \textbf{Exclusions}: Generic facilities (PACU), generic family terms (sister), generic teams (TT CARDIAC ICU), generic services (language line), medical codes (P245).
\end{itemize}

\medskip
\textbf{OUTPUT FORMAT}

\noindent Return valid JSON only. No markdown, explanations, or extra text.

\noindent\texttt{\{``ENTITY\_TYPE'': [\{``text'': ``exact text'', ``confidence'': 0.95\}]\}}

\noindent Entity types: \texttt{PATIENT}, \texttt{LOCATION}, \texttt{PHONE}, \texttt{WEB}, \texttt{DATE}, \texttt{ID}, \texttt{HOSPITAL}, \texttt{DOCTOR}, \texttt{AGE}, \texttt{OTHER}.\\
Confidence: 0.00--1.00 (use 0.90--0.99 for clear PHI). No PHI found: return \texttt{\{\}}. Better to over-tag with lower confidence than miss entities.

\medskip
\noindent\textit{[Two few-shot examples omitted for brevity; available in the released prompt file.]}

\medskip
\noindent\texttt{Clinical Text:}\\
\texttt{\{clinical\_text\}}
\end{tcolorbox}

\section{Cost Analysis: LLM Processing vs.\ Distillation}
\label{appendix:cost-analysis}

This appendix provides the detailed cost derivation referenced in the Discussion. All prices are Gemini 2.5 Flash Flex/Batch tier from Google Cloud Vertex AI \citep{google_vertexai_pricing_2025}, accessed March 2025. Flash has no long-context surcharge: the price is the same regardless of whether input is $\leq$200K or $>$200K tokens.

\subsection{Hypothetical full-warehouse LLM cost}

The STARR-OMOP clinical data warehouse contains approximately \starrChars{} characters of clinical text across all note types. Using a standard rule of thumb of 4 characters per token for clinical English text, this yields an estimated \textbf{\starrTokens{} input tokens}.

The output token estimate is derived from the structured JSON output of the AIMI v2 de-identification pipeline. Each clinical note produces a JSON object containing identified PHI entities with text spans, entity types, and confidence scores. Scaling the observed per-note JSON output to the full warehouse yields an estimated \textbf{536 billion output tokens}. This is a conservative lower-bound estimate; actual output may be larger depending on the density of PHI entities and JSON formatting overhead.

\begin{table}[H]
  \caption{Hypothetical cost of processing the entire STARR-OMOP warehouse through Gemini 2.5 Flash (Flex/Batch pricing, March 2025).}
  \label{tab:cost-warehouse}
  \centering
  \small
  \begin{tabular}{lrrr}
    \toprule
    \textbf{Component} & \textbf{Tokens} & \textbf{Price per 1M} & \textbf{Cost} \\
    \midrule
    Input (clinical text)  & \starrTokensShort{}  & \$0.15 & \$23,738 \\
    Output (JSON entities) & 536B  & \$1.25 & \$670,000 \\
    \midrule
    \textbf{Total}         &       &        & \textbf{\$693,738} \\
    \bottomrule
  \end{tabular}
\end{table}

\subsection{Distillation as a one-time alternative}

Our distillation pipeline queries Gemini 2.5 Flash only once, on \teacherLabelNotes{} notes, to generate silver-standard labels. The resulting DeBERTa v3 model (\studentParams{} parameters) then runs locally on standard CPU or consumer GPU hardware and is not re-billed per note. The recurring cloud cost is therefore incurred a single time on a small fraction of the warehouse, rather than on the full \starrTokensShort{} tokens for every pass. This one-time labeling excludes human annotation effort for the \shieldSelectedNotes{}-note SHIELD gold standard and local compute for student model training, which are fixed costs independent of warehouse size.

\noindent\textbf{Assumptions and caveats.}
\begin{itemize}[leftmargin=*]
    \item Token counts use a 4-character-per-token rule of thumb. Actual tokenization depends on the model's tokenizer (SentencePiece for Gemini) and may vary by $\pm$20\% for clinical text with abbreviations and medical terminology.
    \item Output token estimates are derived from the structured JSON output of the AIMI v2 pipeline and represent a conservative lower bound.
    \item Flex/Batch pricing (\$0.15/1M input, \$1.25/1M output) is the lowest available tier; standard pricing (\$0.30 input, \$2.50 output) would double the cost to $\sim$\$1.39M. Priority pricing (\$0.54 input, \$4.50 output) would increase it further to $\sim$\$2.50M.
    \item The estimate excludes API overhead, network transfer costs, HIPAA-compliant infrastructure, and the operational complexity of processing \starrTokensShort{} tokens through cloud APIs while maintaining data governance compliance.
\end{itemize}

\section{Public Release: Statistics and Demographics}
\label{appendix:release}

This appendix reports descriptive statistics for the released SHIELD corpus (\shieldNotes{} notes after deduplication) referenced in Section~\ref{sec:dataset-release}: PHI category distribution, note-length distribution, patient demographics, and the most common clinical note types. To match the suppression policy of the public release, cells corresponding to fewer than 10 observations are reported as \texttt{(<10)}.

\begin{table}[H]
  \caption{PHI category distribution across the \shieldNumCats{} canonical SHIELD categories.}
  \label{tab:shield-phi-distribution}
  \centering
  \small
  \begin{tabular}{lrrr}
    \toprule
    \textbf{Category} & \textbf{Count} & \textbf{Percentage} & \textbf{Mean per note} \\
    \midrule
    DATE & 3{,}547 & 34.7\% & 2.6 \\
    DOCTOR & 2{,}598 & 25.4\% & 1.9 \\
    PATIENT & 1{,}031 & 10.1\% & 0.7 \\
    HOSPITAL & 922 & 9.0\% & 0.7 \\
    ID & 728 & 7.1\% & 0.5 \\
    PHONE & 513 & 5.0\% & 0.4 \\
    LOCATION & 445 & 4.4\% & 0.3 \\
    AGE & 363 & 3.5\% & 0.3 \\
    WEB & 82 & 0.8\% & 0.1 \\
    \bottomrule
  \end{tabular}
\end{table}

\begin{table}[H]
  \caption{Note-length distribution (characters and words per note).}
  \label{tab:shield-text-stats}
  \centering
  \small
  \begin{tabular}{lrrrrrr}
    \toprule
    \textbf{Metric} & \textbf{Min} & \textbf{Q1} & \textbf{Median} & \textbf{Mean} & \textbf{Q3} & \textbf{Max} \\
    \midrule
    Characters & 300 & 654 & 1{,}015 & 1{,}166 & 1{,}602 & 2{,}539 \\
    Words & 28 & 94 & 146 & 168 & 228 & 438 \\
    \bottomrule
  \end{tabular}
\end{table}

\begin{table}[H]
  \caption{Patient sex distribution (one row per unique patient).}
  \label{tab:shield-demo-sex}
  \centering
  \small
  \begin{tabular}{lrr}
    \toprule
    \textbf{Category} & \textbf{Count} & \textbf{Percentage} \\
    \midrule
    FEMALE & 710 & 51.6\% \\
    MALE & 657 & 47.7\% \\
    No matching concept & (<10) &  \\
    \bottomrule
  \end{tabular}
\end{table}

\begin{table}[H]
  \caption{Patient race distribution (one row per unique patient).}
  \label{tab:shield-demo-race}
  \centering
  \small
  \begin{tabular}{lrr}
    \toprule
    \textbf{Category} & \textbf{Count} & \textbf{Percentage} \\
    \midrule
    No matching concept & 415 & 30.2\% \\
    White & 387 & 28.1\% \\
    Asian & 220 & 16.0\% \\
    Black or African American & 174 & 12.6\% \\
    Native Hawaiian or Other Pacific Islander & 110 & 8.0\% \\
    American Indian or Alaska Native & 70 & 5.1\% \\
    \bottomrule
  \end{tabular}
\end{table}

\begin{table}[H]
  \caption{Patient ethnicity distribution (one row per unique patient).}
  \label{tab:shield-demo-ethnicity}
  \centering
  \small
  \begin{tabular}{lrr}
    \toprule
    \textbf{Category} & \textbf{Count} & \textbf{Percentage} \\
    \midrule
    Not Hispanic or Latino & 756 & 54.9\% \\
    Hispanic or Latino & 401 & 29.1\% \\
    No matching concept & 219 & 15.9\% \\
    \bottomrule
  \end{tabular}
\end{table}

\begin{table}[H]
  \caption{Patient age distribution (one row per unique patient).}
  \label{tab:shield-demo-age}
  \centering
  \small
  \begin{tabular}{lrr}
    \toprule
    \textbf{Age band} & \textbf{Count} & \textbf{Percentage} \\
    \midrule
    65-79 & 340 & 24.7\% \\
    50-64 & 260 & 18.9\% \\
    35-49 & 219 & 15.9\% \\
    18-34 & 201 & 14.6\% \\
    80+ & 197 & 14.3\% \\
    0-17 & 159 & 11.6\% \\
    \bottomrule
  \end{tabular}
\end{table}

\begin{table}[H]
  \caption{Top 10 clinical note types by frequency.}
  \label{tab:shield-note-types}
  \centering
  \small
  \begin{tabular}{lrr}
    \toprule
    \textbf{Note type} & \textbf{Count} & \textbf{Percentage} \\
    \midrule
    letter & 45 & 3.3\% \\
    imaging & 36 & 2.6\% \\
    procedures & 36 & 2.6\% \\
    progress notes & 34 & 2.5\% \\
    clinic support note & 32 & 2.3\% \\
    telephone encounter & 31 & 2.2\% \\
    discharge instructions & 31 & 2.2\% \\
    patient instructions & 31 & 2.2\% \\
    nursing note & 29 & 2.1\% \\
    consults & 27 & 2.0\% \\
    \bottomrule
  \end{tabular}
\end{table}

\end{document}